\pdfoutput=1

\documentclass[11pt]{article}

\usepackage[preprint]{acl}

\usepackage{times}
\usepackage{latexsym}

\usepackage[T1]{fontenc}

\usepackage[utf8]{inputenc}

\usepackage{microtype}

\usepackage{inconsolata}

\usepackage{graphicx}

\usepackage{makecell}
\usepackage{adjustbox}
\usepackage{amsmath}
\usepackage{amssymb}
\usepackage{amsthm}
\usepackage[linesnumbered,ruled,vlined]{algorithm2e}
\SetAlgoNlRelativeSize{-1}
\usepackage{float}
\usepackage{xspace}
\usepackage{booktabs}
\usepackage{siunitx}
\usepackage{multirow}
\usepackage{colortbl}
\definecolor{McColor}{rgb}{0.53,0.88,0.91}
\definecolor{DcColor}{rgb}{0.76,0.71,1.00}
\usepackage{placeins}
\usepackage{enumitem}

\newcommand{\Dense}{\texttt{Dense}\xspace}
\newcommand{\Spideal}{\texttt{SP\textsuperscript{Ideal}}\xspace}
\newcommand{\Spprac}{\texttt{SP\textsuperscript{Prac}}\xspace}
\newcommand{\Idx}{\texttt{$I\!D\!X$}\xspace}
\newcommand{\Thld}{\texttt{$T\mkern-2mu H\mkern-2mu L\mkern-2mu D$}\xspace}
\newcommand{\COUNTDOWN}{\textsc{CountDown}\xspace}
\newcommand{\MC}{\textsc{M-CountDown}\xspace}
\newcommand{\DC}{\textsc{D-CountDown}\xspace}
\newcommand{\MCFULL}{\textsc{Mono-CountDown}\xspace}
\newcommand{\DCFULL}{\textsc{Dual-CountDown}\xspace}

\title{COUNTDOWN: Contextually Sparse Activation Filtering Out\protect\\ Unnecessary Weights in Down Projection}

\author{Jaewon Cheon \\
  Industrial and Management Engineering \\
  Korea University \\
  \texttt{jaewon\_cheon@korea{.}ac{.}kr} \\\And
  Pilsung Kang* \\
   Industrial Engineering \\
   Seoul National University \\
  \texttt{pilsung\_kang@snu{.}ac{.}kr} \\}

\begin{document}
\maketitle
\begin{abstract}
The growing size of large language models has created significant computational inefficiencies. To address this challenge, sparse activation methods selectively deactivate non-essential parameters during inference, reducing computational costs in Feed-Forward Networks (FFN) layers. While existing methods focus on non-linear gating mechanisms, we hypothesize that the sparsity lies globally in the form of a linear combination over its internal down projection matrix. Based on this insight, we propose two methods: \MC, leveraging indirect coefficients, and \DC, utilizing direct coefficients of the linear combination. Experimental results demonstrate that \DC can omit 90\% of computations with performance loss as low as 5.5\% ideally, while \MC provides a predictor-free solution with up to 29.4\% better performance preservation compared to existing methods. Our specialized kernel implementations effectively realize these theoretical gains into substantial real-world acceleration. We release our code at \href{https://github.com/rustic-snob/COUNTDOWN}{COUNTDOWN}.
\end{abstract}

\section{Introduction}

Large Language Models (LLMs) have demonstrated remarkable capabilities across diverse applications, from handling specific tasks to orchestrating agent-based operations \citep{OpenAI2024-ak, DeepSeek-AI2024-ff, Gemma_Team2025-ak}. However, these advancements came at the cost of dramatically increased model sizes, creating enormous computational and resource demands.

The inference process has emerged as a particularly acute efficiency constraint, forming a critical bottleneck for deploying LLMs in practical applications. This inefficiency is further amplified by recent trends in test-time scaling, where models generate extensive reasoning, significantly increasing computational demands during inference \citep{Jang2024-pg, Deng2024-it}. Consequently, research on LLM inference efficiency has intensified, aiming to reduce latency and memory consumption while preserving generation quality \citep{Liu2024-ee, Kwon2023-ye, Cai2024-dk}.
\begin{figure}[t]
    \centering
    \includegraphics[width=\linewidth]{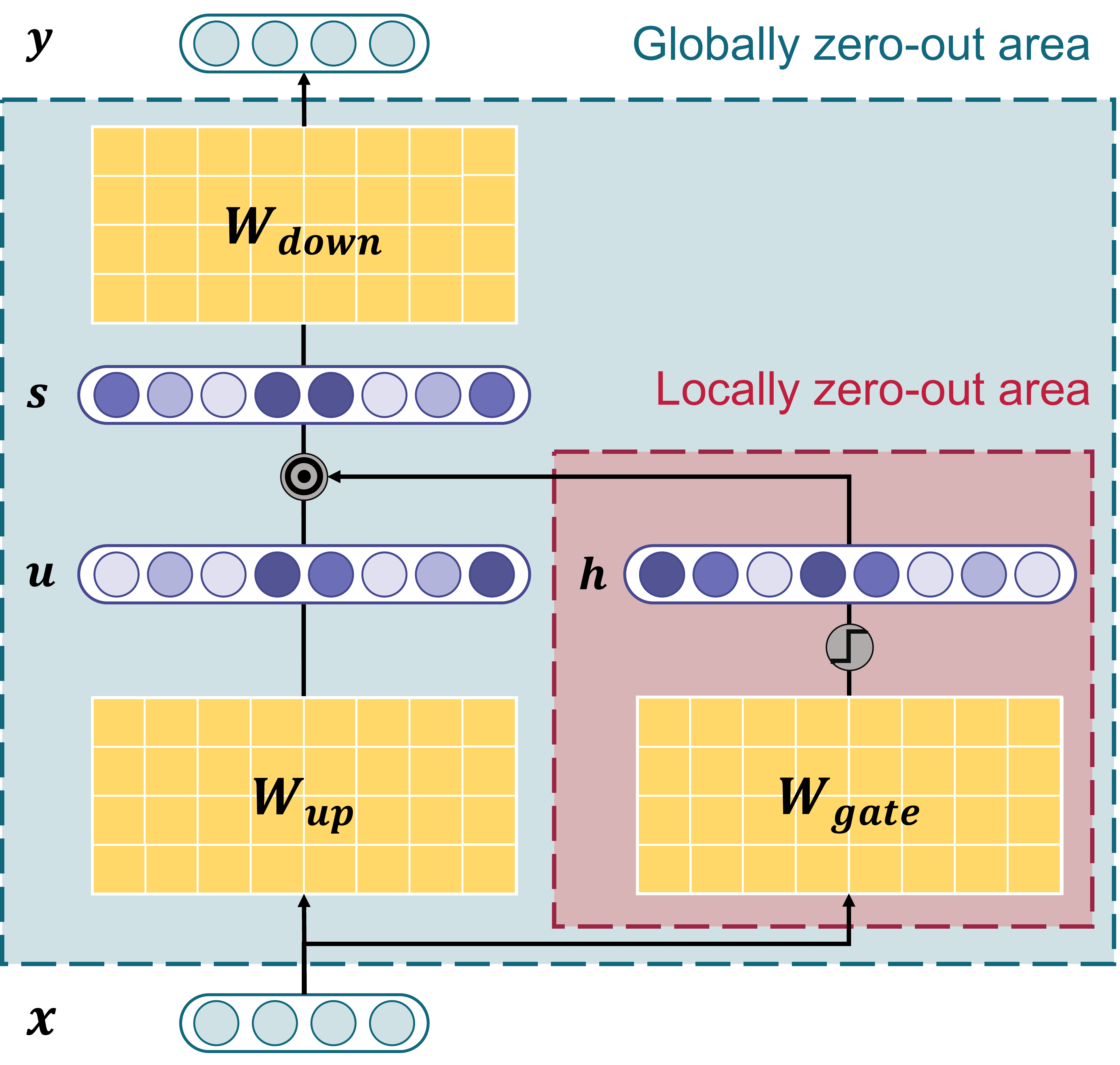}
    \vspace{-1.5em}
    \caption{Comparison of sparsity determinations: our approach determines sparsity from the full FFN computation (turquoise box), whereas conventional methods like CATS \citep{Lee2024-zb} rely solely on non-linear activations (red box).}
    \label{fig:coverage}
    \vspace{-1.4em}
\end{figure}

In this context, sparse activation has emerged as a prominent strategy to improve the efficiency of the Feed-Forward Networks (FFN) in LLM \citep{Liu2023-cn, Lee2024-zb, Akhauri2024-dk, Alizadeh2024-yu}. Sparse activation methods dynamically identify and deactivate parameters unnecessary for a given input, thereby reducing computational load and accelerating inference. These methods are particularly beneficial since FFN layers incur significant computational overhead in modern LLM architectures \citep{Awasthi2024-mc}.

The zero-out gating property of ReLU \citep{agarap2019deeplearningusingrectified} creates extensive sparsity in FFN layers by forcing a large portion of neurons to output zero \citep{Mirzadeh2024-zh}. This natural sparsity makes computations associated with these zero-valued neurons completely redundant. Existing sparse activation methods leverage this property to identify and skip these unnecessary computations \citep{Sun2024-el, Zhang2024-as}. However, recent LLMs largely employ activations such as GeLU or SiLU \citep{Hendrycks2016-kl, Elfwing2018-ww} with far less prevalent zero-out behavior \citep{Mirzadeh2024-zh}, limiting these methods’ applicability. Further, Gated-MLP structures, now widely adopted as FFN layers \citep{Shazeer2020-zg, pmlr-v70-dauphin17a}, introduce more complex parameter interactions than standard architectures. This invalidates the assumption that sparsity occurs only around non-linear activations.

To overcome these limitations, we propose an approach that defines sparsity from a global view, extending beyond the non-linear activations by reformulating the FFN layer's output as a weighted sum, as illustrated in \autoref{fig:coverage}. Based on this approach, we derive two sparse activation methodologies: \MCFULL (\MC) and \DCFULL (\DC). \MC identifies sparsity based on the output of a single weight matrix in Gated-MLP, while \DC leverages two weight matrices. In evaluations, \MC consistently outperforms the baseline method CATS \citep{Lee2024-zb}, achieving up to 29.4\% better performance preservation with comparable inference speed. \DC attains greater efficiency gains, reducing computations by up to 90\% in FFN layers with performance loss as low as 5.5\% under optimal conditions.

The contributions of this paper are as follows.
\begin{itemize}[topsep=0.5em, partopsep=0pt, itemsep=0.2em, parsep=0pt]
  \item We introduce a novel theoretical framework that redefines sparsity through a weighted-sum perspective over down projection matrices, extending beyond the conventional focus on activation functions.
  \item We demonstrate that analyzing coefficient vectors in the weighted sum enables superior sparsity decisions, resulting in two distinct approaches with complementary strengths.
  \item We provide practical acceleration through optimized kernel implementations, enabling both methods to achieve substantial throughput improvements across multiple state-of-the-art LLM architectures.
\end{itemize}

\section{Related Works}

\paragraph{ReLU-based Sparse Activation}  
Early works on sparse activation primarily leveraged the property of ReLU to enhance computational efficiency. These approaches identified that ReLU activation functions naturally create substantial built-in sparsity by producing zeros for negative values \citep{DBLP:conf/iclr/LiYBLRRYCYGK23}. Several approaches have tried to detect these zero-valued activations to preemptively skip associated computations, as these neurons would have no impact on subsequent layers \citep{Mirzadeh2024-zh}. Deja Vu \citep{Liu2023-cn} extended this concept by training lightweight predictors to anticipate which neurons would be zeroed out, further improving efficiency. While these methods showed impressive speed gains with minimal performance loss, their application faced significant constraints. Notably, these approaches were practical only on architectures explicitly designed with ReLU activations, limiting their applicability as LLMs increasingly adopted alternative activation functions \citep{Akhauri2024-dk}.

\paragraph{Non-ReLU Sparse Activation}  
As LLMs evolved to favor non-ReLU activation functions such as GeLU and SiLU, which rarely produce exact zeros, new methods emerged to extend sparsity benefits to these architectures. One direction involved ReLUfication techniques that replace non-ReLU functions with ReLU, enabling the reuse of existing sparsity strategies \citep{Song2024-rv, Song2025-zh, Zhang2024-as, Alizadeh2024-yu}. Another approach, such as by CATS \citep{Lee2024-zb}, redesigned sparsity criteria to identify and skip computations associated with near-zero activations rather than exact zeros. While these adaptations improved compatibility with modern LLM architectures, they remain fundamentally constrained by their narrow focus on local patterns around non-linear transformations, overlooking potential sparsity from a global perspective of the FFN layer. This localized perspective may fail to fully capitalize on the potential sparsity distributed throughout modern Gated-MLP architectures, particularly considering the complex interactions among multiple weight matrices that define these structures.

\section{Generalization of Sparse Activation}

\paragraph{Problem Formulation}

A Gated-MLP block consists of three weight matrices: $W_{up},$ $W_{gate},$ $W_{down}$ $\in\mathbb{R}^{d_{\text{model}}\times d_{\text{inter}}}$. For this block, the input vector $x$ and the output vector $y$ are in $\mathbb{R}^{d_{\text{model}}}$. The computation involves intermediate states defined as $u\mkern-2mu =\mkern-2mu x\mkern-2mu \cdot\mkern-2mu W_{up},$ $h\mkern-2mu =\mkern-2mu \sigma(x\mkern-2mu \cdot \mkern-2mu W_{gate}),$ $s\mkern-2mu =\mkern-2mu u\mkern-2mu \odot\mkern-2mu h$ in $\mathbb{R}^{d_{\text{inter}}}$.

When no sparsification is applied, which we refer to as the \Dense scenario, all parameters are activated, and the operation proceeds as follows:
\begin{equation}
  y = \bigl((x\cdot W_{up}) \odot \sigma (x\cdot W_{gate})\bigr)\cdot W^\intercal_{down}
  \label{eq:dense_computation}
\end{equation}
where $\sigma$ denotes a non-linear activation function, typically GeLU or SiLU.

We now introduce our sparsity propagation framework, establishing sparse activation from a global perspective. We can activate only a valuable subset of weight vectors, with a marginal performance loss. Specifically, sparse activation under our framework follows:
\begin{equation}
  y = \bigl((x\cdot W_{up}^{I})\odot\sigma(x\cdot W_{gate}^{I})\bigr)\cdot \bigl(W^{I}_{down}\bigr)^\intercal
  \label{eq:sparse_simple}
\end{equation}
where $I$ denotes the column of indices of the weights selected for computation:
\begin{equation}
  W^{I} = W[:,\Idx]\,,\quad\Idx = \Thld(\cdot)
  \label{eq:index_definition}
\end{equation}
where \Thld is any function filtering effective $I$.

Notably, when individual threshold functions are defined separately for each matrix, identical output can be achieved through the unified intersection \Idx:
\begin{equation}
    \Idx = \Idx_{up}\cap\Idx_{gate}\cap\Idx_{down}
    \label{eq:shared-index behavior}
\end{equation}

Consequently, even when sparsifying just one matrix and keeping others dense, the computation remains equivalent to applying this unified $\Idx$ across all matrices, which we denotes as shared-index property. Thus, if valuable sparsity patterns are identified in one matrix, they can propagate throughout the entire Gated-MLP.

A critical challenge, therefore, is defining the optimal filtering function \Thld to identify the most effective index set \Idx to preserve globally essential computations while significantly reduce computational overhead.

\paragraph{Limitation of Comparative Methodology} 

CATS \citep{Lee2024-zb} partially satisfies our sparsity propagation framework. It identifies sparsity by examining the activation magnitude $h = \sigma(x\mkern-2mu \cdot\mkern-2mu W_{gate})$, assuming activations squashed near zero indicate parameters to omit. Specifically, given a sparsity ratio $k \in (0,1)$, CATS computes a threshold $\tau_{\text{C}}^k$ via the $\text{Quantile}(k, |h|)$ operation, selecting a cutoff below which the lowest $k$ fraction of activations is excluded. Based on this threshold, CATS defines a sparse activation index as shown in \autoref{eq:sparse_cats}.

CATS leverages the shared-index property. However, since the optimal $\Thld$ might depend on factors beyond non-linear activation region, CATS is theoretically limited in propagating an optimal $\Idx$ throughout the Gated-MLP. Additionally, although $h[i]$ is large, if the corresponding $u[i] = x \cdot W_{up}[i]$ is near zero, the final contributions become minimal, which ideally should be filtered out due to their elementwise product. 


\paragraph{Threshold Variants}  
To overcome these limitations, we reformulate the Gated-MLP computation as a linear combination of the $W_{down}$ weight vectors, thereby exploring additional possibilities for defining $\Thld$ as follows:
\vspace{-0.5em}

\begin{equation}
\begin{aligned}
  y =& \bigl((x\cdot W_{up}) \odot \sigma (x\cdot W_{gate})\bigr)\cdot W^\intercal_{down} \\
  =& \sum_i s[i] \cdot W^\intercal_{down}[i]
\end{aligned}
\label{eq:linear_combi}
\end{equation}

This reformulation allows us to interpret output $y$ as a weighted sum over $W^\intercal_{down}$ row vectors, where coefficient $s[i] = \bigl((x \cdot W_{up}) \odot \sigma(x \cdot W_{gate})\bigr)[i]$ reflects the $i$-th row vector's contribution to computation. The magnitude of these coefficients provides a natural metric for determining which parameters to activate, as they quantify each vector's significance to the output.

Furthermore,  since $s$ is calculated as the elementwise multiplication of $u = x \cdot W_{up}$ and $h = \sigma(x \cdot W_{gate})$, these intermediate vectors can also serve as indirect coefficient signals. This generalized view reveals that each computation stage in the Gated-MLP can provide a distinct sparsity indicator, with selecting $h$ as the basis being equivalent to CATS's approach.
\vspace{-1em}

\begin{subequations}
\label{eq:sparse_categorized}
\begin{align}
\Thld_{\text{C}}^k\,(h,\tau_{\text{C}}^k) &= \{ i \mid \lvert h[i] \rvert > \tau_{\text{C}}^k \} \label{eq:sparse_cats} \\ 
\Thld_{\text{M}}^k\,(u,\tau_{\text{M}}^k) &= \{ i \mid \lvert u[i] \rvert > \tau_{\text{M}}^k \} \label{eq:sparse_countdown_m} \\ 
\Thld_{\text{D}}^k\,(s,\tau_{\text{D}}^k) &= \{ i \mid \lvert s[i] \rvert > \tau_{\text{D}}^k \} \label{eq:sparse_countdown_d}
\end{align}
\end{subequations}

Based on this view, we propose two variants of sparse activation that extend beyond prior approaches relying solely on the magnitude of $h$. As shown in \autoref{eq:sparse_categorized}, where subscripts C, M, and D denote CATS, \MC, and \DC methods respectively, the first method, \MC, applies thresholding directly to vector $u$, while the second method, \DC, applies thresholding to $s$. For each method, thresholds $\tau_{\text{M}}^k$ and $\tau_{\text{D}}^k$ are calculated via $\text{Quantile}(k, |u|)$ and $\text{Quantile}(k, |s|)$ respectively.

These methods offer complementary strengths: \MC provides practical implementation with minimal overhead by examining only one matrix multiplication, while \DC can offer more precise sparsity identification through direct coefficients of the weighted sum. A detailed discussion of these methods follows in \autoref{sec:realization}

\begin{figure*}[tb]
    \centering
    \includegraphics[width=\textwidth]{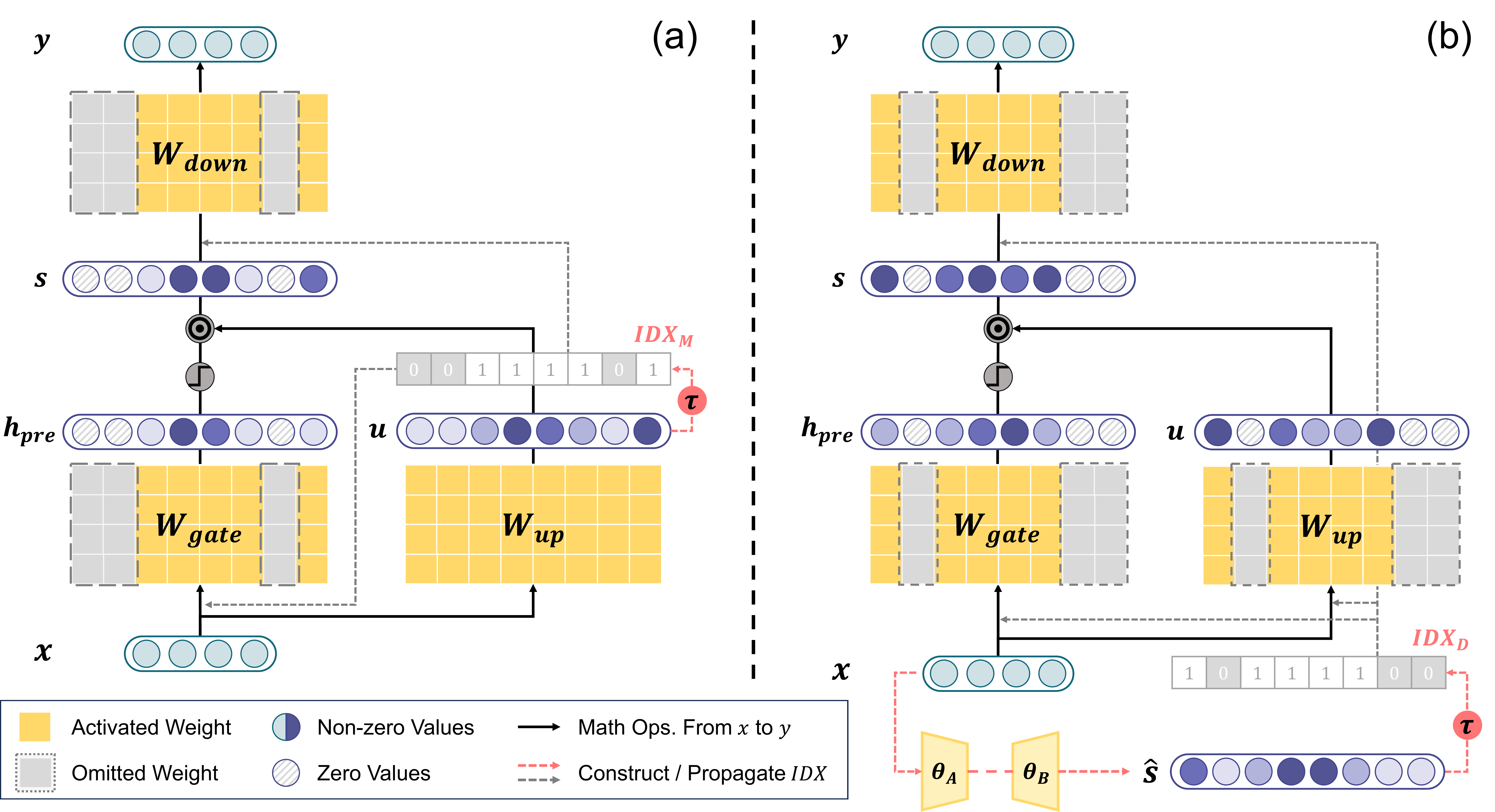}
    \caption{\COUNTDOWN Pipeline. Note that $h_{pre}=x\cdot W_{gate}$. \emph{Left (a):} In \MC, we determine which parameters to activate by binarizing densely computed $u$ with pre-calculated $\hat{\tau_\text{M}^k}$. \emph{Right (b):} In \DC, low-rank predictors $\bigl(\theta_\text{A}$, $\theta_\text{B}\bigr)$ determine which parameters to activate.}
    \vspace{-1em}
    \label{fig:countdown}
\end{figure*}

\section{Implementation of Sparse Activation}
\label{sec:realization}

\paragraph{\Spideal\space and\space\Spprac}
\label{para:sp_ideal_and_sp_prac}

In the previous section, we focused on establishing \Thld and the corresponding indicator that theoretically guarantee the safe omission of parameters. Ideally, if these indicators are tractable in real-time inference, we can achieve the upper-bound performance defined by the method. However, accessing the indicator and deriving \Idx from it is not trivial.

Given this constraint, we distinguish between two distinctive perspectives: \textbf{\Spideal} examines the theoretical upperbound performance achievable by each method, assuming that filtering based on sparsity indicators incurs no computational overhead. \textbf{\Spprac} accounts for real-world deployment constraints, particularly the latency of identifying sparse activation patterns. It evaluates whether methods can deliver actual inference speedups when all practical overheads are considered.

The distinction is critical because methods with strong \Spideal performance may not translate to \Spprac benefits if their practical implementation is computationally expensive. Conversely, focusing solely on \Spprac without understanding the theoretical \Spideal limits can lead to suboptimal solutions that fail to approach the best possible performance. Effective sparse activation requires both identifying truly essential computations via \Spideal and creating an efficient implementation to realize total computational savings through \Spprac.

\paragraph{Constructing \Spprac for \COUNTDOWN}

We now describe how to transform the theoretical \Spideal formulations of \MC and \DC into efficient, practical \Spprac implementations.

For \MC, the implementation is straightforward because its indicator $u$ depends only on the matrix $W_{up}$. Therefore, its index set $\Idx_\text{M}^k$ defined in its \Spideal perspective can be obtained independently of other matrices in the Gated-MLP. This allows \MC to operate without additional inference-time components, as computation over the remaining matrices can be selectively skipped based on $u$.

To further reduce overhead, we avoid computing $\tau_\text{M}^k$ dynamically for each input. Instead, we approximate it with a layerwise constant $\hat{\tau}_\text{M}^k = \frac{1}{T} \sum_{t=1}^T \text{Quantile}(k, |u^{(t)}|)$ estimated during a calibration phase with $T$ sampled inputs.

In contrast, implementing \DC poses greater challenges because its indicator $s$ requires nearly the entire Gated-MLP computation, negating the advantages of sparse activation. To tackle this challenge, we train a lightweight predictor that estimates the optimal index set $\Idx_\text{D}^k$ directly from input $x$, avoiding the need to compute $s$ during inference. For each layer, the predictor outputs a score vector $\hat{s}$ where:
\vspace{-0.2em}
\[
\hat{s}[i] =
\begin{cases}
+\infty & \text{if } |s[i]| > \text{Quantile}(k, |s|) \\
-\infty & \text{otherwise}
\end{cases}
\]
\vspace{-0.2em}

Using this output, we define the predicted index set as $\widehat{\Idx}_\text{D}^k = \{ i \mid \hat{s}[i] > 0 \}$ and activate only the corresponding weight columns during inference.

For efficiency, the predictor must be highly accurate and computationally inexpensive during inference. Following \citep{Liu2023-cn, Alizadeh2024-yu}, we employ a low-rank approximator consisting of two matrices: $\theta_\text{A} \in \mathbb{R}^{d_{\text{model}} \times d_{\text{rank}}}$ and $\theta_\text{B} \in \mathbb{R}^{d_{\text{rank}} \times d_{\text{inter}}}$, minimizing computational overhead while preserving prediction accuracy. \autoref{algo:predictrain} details the complete training procedure.

\paragraph{Hardware Aware Kernel Design}

Once the sparse activation index set $\Idx$ is determined, computation can be restricted to only the corresponding subset of weights, reducing the actual floating-point operation count (FLOPs). However, reducing FLOPs does not necessarily translate to improved inference latency. For instance, materializing an indexed weight matrix and performing standard vector-matrix multiplication may still reduce FLOPs, but at the cost of increased memory access \citep{DBLP:conf/sosp/SongMX024, xue2024powerinfer}. Therefore, sparse computation should avoid incurring excessive memory traffic solely for the sake of reducing arithmetic operations.

To address this, we implement custom kernels for both \MC and \DC using Triton \citep{DBLP:conf/pldi/TilletKC19}. The \MC kernel builds upon CATS's structure \citep{Lee2024-zb}, but optimizes it by fusing the non-linear activation to reduce additional memory access. For \DC, we design a kernel that efficiently supports predictor-based activation patterns. A naive implementation would require eight separate kernel launches for sparse computation: indexing and GEMV for each of the three matrices, plus non-linear activation and elementwise multiplication. Our implementation compresses this workload into just two kernels. This design ensures that FLOPs reductions directly translate into improved token throughput. Full implementation details and pseudocode are  in \autoref{algo:mcountdown_kernel} and \autoref{algo:dcountdown_kernel}.

\begin{table*}[t]
  \centering
  \resizebox {\textwidth} {!}{
  \begin{tabular}{l|ccc|ccc|ccc|ccc}
    \toprule
    \multirow{2}{*}{\bf InferenceMode}
      & \multicolumn{3}{c}{\bf Llama-3.1-8B-Instruct} 
      & \multicolumn{3}{c}{\bf gemma-2-9b-it} 
      & \multicolumn{3}{c}{\bf Qwen2.5-14B-Instruct}
      & \multicolumn{3}{c}{\bf phi-4}\\
    \cmidrule(lr){2-4}\cmidrule(lr){5-7}\cmidrule(lr){8-10}\cmidrule(lr){11-13}
      &  $k\!=\!0.7$ &  $k\!=\!0.8$ &  $k\!=\!0.9$ 
      &  $k\!=\!0.7$ &  $k\!=\!0.8$ &  $k\!=\!0.9$ 
      &  $k\!=\!0.7$ &  $k\!=\!0.8$ &  $k\!=\!0.9$
      &  $k\!=\!0.7$ &  $k\!=\!0.8$ &  $k\!=\!0.9$ \\
    \midrule
    \multicolumn{13}{l}{\textbf{Dense}} \\
    \multicolumn{1}{l|}{Full} 
      & \multicolumn{3}{c|}{\textbf{0.616}}
      & \multicolumn{3}{c|}{\textbf{0.645}}
      & \multicolumn{3}{c|}{\textbf{0.674}}
      & \multicolumn{3}{c}{\textbf{0.655}} \\
    \midrule
    \multicolumn{13}{l}{\textbf{\Spideal}} \\
    DEJAVU 
      & 0.314 & 0.315 & 0.322
      & 0.360 & 0.360 & 0.360 
      & 0.379 & 0.382 & 0.385
      & 0.398 & 0.405 & 0.396 \\
    CATS   
      & 0.471 & 0.412 & 0.337
      & 0.592 & 0.483 & 0.367 
      & 0.502 & 0.428 & 0.389
      & 0.615 & 0.535 & 0.427 \\
    \MC    
      & 0.570 & 0.513 & 0.421
      & 0.624 & 0.607 & 0.549 
      & 0.644 & 0.610 & 0.479
      & 0.636 & 0.608 & 0.512 \\
    \DC    
      & \textbf{0.603} & \textbf{0.587} & \textbf{0.525}
      & \textbf{0.635} & \textbf{0.625} & \textbf{0.590} 
      & \textbf{0.660} & \textbf{0.647} & \textbf{0.555}
      & \textbf{0.651} & \textbf{0.649} & \textbf{0.620} \\
    \midrule
    \multicolumn{13}{l}{\textbf{\Spprac}} \\
    CATS   
      & 0.504 & 0.450 & 0.350
      & 0.605 & 0.502 & 0.360 
      & 0.556 & 0.478 & 0.390
      & 0.633 & 0.591 & 0.448 \\
    \MC    
      & \textbf{0.571} & \textbf{0.528} & \textbf{0.447}
      & \textbf{0.632} & \textbf{0.617} & \textbf{0.588} 
      & \textbf{0.651} & \textbf{0.624} & \textbf{0.535}
      & \textbf{0.639} & \textbf{0.620} & \textbf{0.555} \\
    \DC    
      & 0.442 & 0.419 & 0.387
      & 0.555 & 0.563 & 0.520 
      & 0.526 & 0.457 & 0.437
      & 0.499 & 0.445 & 0.417 \\
    \bottomrule
  \end{tabular}}
  \caption{Average \Spideal and \Spprac scores compared to \Dense across all downstream tasks. Full task‐wise results are provided in Appendix~\ref{sec:full_result}.}
  \label{tab:avg_dstask_main}
  \vspace{-1em}
\end{table*}

\section{Experiments}
\label{sec:Experiments}

\paragraph{Experimental Setup}

We evaluate the proposed methods against other sparse activation baselines, primarily CATS \citep{Lee2024-zb} and Deja Vu \citep{Liu2023-cn}. We also include a \Dense variant without any sparse activation for comparison. Experiments are conducted using four diverse state-of-the-art LLMs ranging from 8B to 14B parameters: Llama-3.1-8B-Instruct \citep{Grattafiori2024-zf}, gemma-2-9b-it \citep{Gemma-Team2024-bh}, Qwen2.5-14B-Instruct \citep{Qwen2024-dq}, and phi-4 \citep{abdin2024phi4technicalreport}. We test multiple sparsity ratios by varying $k$ from 0.7 to 0.9, representing the fraction of parameters excluded from computation. Implementation details are provided in Appendix~\ref{para:exp_env}.

We examine both model performance preservation and computational efficiency. For model performance, we use the lm-eval-harness \citep{eval-harness} framework to assess downstream tasks including ARC \citep{Clark2018-sw}, HellaSwag \citep{Zellers2019-or}, PIQA \citep{Bisk2020-la}, OpenbookQA \citep{Mihaylov2018-sl}, TruthfulQA \citep{Lin2022-ku}, WinoGrande \citep{Sakaguchi2020-wv}, and GSM8K \citep{Cobbe2021-ou}. Unlike prior sparse activation studies, we also evaluate conversational ability using LLM-as-a-Judge framework AlpacaEval 2.0~\citep{alpaca_eval}.

To assess computational efficiency and inference speed, we benchmark kernel-level latency to quantify Gated-MLP speedups from sparse activation. We also measure end-to-end token throughput and analyze theoretical reductions in floating-point operations (FLOPs) and memory traffic.

\paragraph{Downstream Task Performance}

As shown in \autoref{tab:avg_dstask_main}, in the \Spideal setting, \DC consistently outperforms all methods across all models and sparsity ratios, exhibiting negligible degradation even when compared to the dense baseline. This demonstrates the effectiveness of \DC's sparsity criterion: the indicator $s$ accurately reflects each parameter's importance to the final output, serving as the coefficient in our linear combination formulation. This provides more informed filtering than methods like CATS which rely solely on gating magnitude. Even at 90\% sparsity, \DC retains only the most impactful neurons, limiting performance drop to 5.5\% in the best case among evaluated models.

\MC,  although less effective than \DC, consistently outperforms CATS. The gap between the two widens as the sparsity ratio increases, reaching over 29.4\%. This demonstrates that \MC’s indicator $u$ is more predictive of useful computation than CATS’ indicator $h$. This may seem counterintuitive since $u$ and $h$ contribute symmetrically via their elementwise product and thus should be equally informative. We revisit this comparison in \autoref{para:uvsh}.

Deja Vu, which assumes ReLU-style zero-out behavior, suffers severe degradation in the \Spideal setting. Given its reliance on predictors, which would further degrade under the \Spprac setting, we excluded it from subsequent experiments.

In the \Spprac setting, \DC experiences performance loss relative to the \Spideal due to predictor sub-optimality, suggesting better prediction strategies are needed to fully realize its potential in deployment. In contrast, \MC, thanks to its predictor-free design, exhibits nearly identical performance to its \Spideal counterpart. Notably, \MC continues to outperform CATS across all sparsity settings, reinforcing the effectiveness of its signal even under realistic constraints.

\paragraph{LLM Chat Performance}
While prior studies rely on downstream task accuracy or perplexity, these metrics often fail to capture conversational performance. To address this, we evaluate each method using an LLM-as-a-Judge framework that directly assesses chat-level performance.

As shown in \autoref{tab:avg_chat_perform}, \MC maintains nearly identical performance between the \Spideal and \Spprac settings, while also outperforming CATS in both. \DC exhibits noticeable degradation in \Spprac due to predictor limitations, but retains a dominant lead under \Spideal. This trend aligns with the results observed in the downstream task evaluations.

\begin{table}[h]
  \centering
  \resizebox {0.9\linewidth} {!}{
  \begin{tabular}{l|c|c|c}
    \toprule
    \multirow{2}{*}{\makecell[{{l}}]{\bf InferenceMode}}
      & \multicolumn{3}{c}{\bf AlpacaEval 2.0} \\
      \cmidrule{2-4}
      & \bf $k\,=\,0.7$
      & \bf $k\,=\,0.8$ 
      & \bf $k\,=\,0.9$\\
    \midrule
    \multicolumn{4}{l}{\textbf{\Spideal}} \\
    CATS   
      & 25.10 & 1.72 & 0.19 \\
    \MC    
      & 45.84 & 29.22 & 3.90 \\
    \DC    
      & \textbf{48.86} & \textbf{45.85} & \textbf{29.95} \\
    \midrule
    \multicolumn{4}{l}{\textbf{\Spprac}} \\
    CATS   
      & 31.63 & 10.47 & 0.25 \\
    \MC    
      & \textbf{38.31} & \textbf{33.80} & \textbf{15.88} \\
    \DC    
      & 3.40 & 2.81 & 1.16 \\
    \bottomrule
  \end{tabular}}
  \caption{Average \Spideal and \Spprac win rates against \Dense across all models. Full model‐wise results are provided in \autoref{tab:win_sparsity}.}
  \vspace{-0.4em}
  \label{tab:avg_chat_perform}
\end{table}

\begin{figure}[t]
    \centering
    \includegraphics[width=\linewidth]{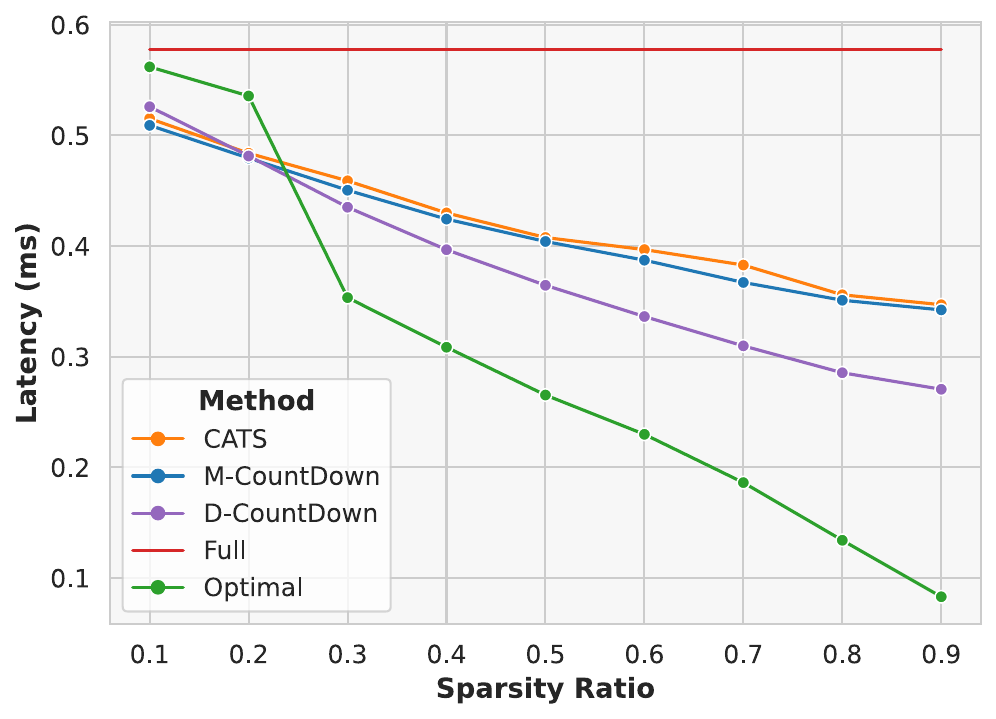}
    \vspace{-1em}
    \caption{Kernel Speed for Llama-3.1-8B-Instruct. CATS, \MC and \DC show their respective \Spprac kernel speeds, Full and Optimal show \Dense while $int({d_{inter}} \times k)$ instead of $d_{inter}$ for the Optimal. Results for other models are in \autoref{fig:apdx_kernel}.}
    \vspace{-1em}
    \label{fig:kernel_speed}
\end{figure}
\paragraph{Efficiency and Speed}
To confirm that reductions in computation indeed translate into inference speedups, we measured kernel-level execution latency under various sparsity ratios. Each kernel's execution time was recorded from the start of the Gated-MLP computation, explicitly excluding other operations like token embedding or attention mechanisms. This allowed us to isolate the precise efficiency gains attributable to sparse activation.

As shown in \autoref{fig:kernel_speed}, \DC achieves the fastest kernel execution time overall, despite the presence of a predictor, by skipping all three weight matrix computations. Although both \MC and CATS are predictor-free, \MC slightly outperforms CATS in kernel speed. Given that the only architectural difference between their kernels is whether the non-linear activation function is fused, this suggests that \MC gains a minor but consistent speed advantage by fusing the activation computation, thereby reducing memory traffic and avoiding additional overhead.

Furthermore, we measured average tokens generated per second for generation lengths of 512 and 1024, providing a model-level speedup assessment in typical generation scenarios. As shown in \autoref{tab:flops_mem_sparsity_time}, \MC achieves the highest end-to-end token throughput. Meanwhile, \DC demonstrates the best performance at the kernel level, and with further optimization, its overall throughput may be further enhanced.

\section{Analysis}
\paragraph{\MC vs CATS}
\label{para:uvsh}

While CATS and \MC share similar core ideas for sparse activation, our experimental results show that \MC consistently achieves better performance. To understand the performance gap between the indirect coefficient vectors $u$ and $h$, we conduct a comparative analysis of how each influences and aligns with the oracle-like reference signal $s$, the direct coefficient used in \DC.

To enable direct comparison, we define binary masks $S^k$, $U^k$, and $H^k$ based on the top-$k$ magnitude entries of each vector. Each mask marks components as “alive” (1) if they survive quantile thresholding, and "dead" (0) otherwise. These binary masks are equivalent to the index sets $\Idx^k$ used for sparse activation, as each represents the support of the corresponding $\Idx^k$ in vector form.

\begin{table}[t]
\centering
\resizebox{\linewidth}{!}{
\begin{tabular}{c|c|cc|>{\centering\arraybackslash}p{1cm}>{\centering\arraybackslash}p{1cm}}
\toprule
\multirow{2}{*}{\textbf{$\bf k$}} 
  & \multirow{2}{*}{\textbf{Method}} 
      & \multirow{2}{*}{\textbf{FLOPs(M)}} 
          & \multirow{2}{*}{\textbf{Mem.(MB)}} 
              & \multicolumn{2}{c}{\textbf{Throughput}} \\
\cmidrule(lr){5-6}
  &       &        &         & $512$ & $1024$  \\
\midrule
0.0 & \Dense & 352.41 & 168.121 
    & 24.64 & 22.63 
 \\
\midrule
\multirow{3}{*}{0.7} 
    & CATS     & 188.00	& 89.746
        & 32.62 & 29.40 
 \\
    & MC       & 187.95	& 89.719 
        & 33.61 & 30.32  \\
    & DC       & \textbf{124.59}	& \textbf{59.480} 
        & \textbf{30.69}  & \textbf{27.80} 
 \\
\midrule
\multirow{3}{*}{0.8} 
    & CATS     & 164.52	& 78.550 
        & 32.72  & 29.60
 \\
    & MC       & 164.46	& 78.522
        & 33.80 & 30.61 
 \\
    & DC       &  \textbf{89.37}	& \textbf{42.684}
        & \textbf{30.70}  & \textbf{27.57} 
 \\
\midrule
\multirow{3}{*}{0.9} 
    & CATS     & 141.02	& 67.345
        & 32.98  & 29.81 
 \\
    & MC       & 140.96 & 67.318 
        & 33.51  & 30.78  \\
    & DC       &  \textbf{54.11}	& \textbf{25.877} 
        & \textbf{30.73}  & \textbf{27.55} 
 \\
\bottomrule
\end{tabular}
}
\caption{Theoretical FLOPs and Memory Traffic of Gated-MLP and actual throughput per second at sequence lengths 512 and 1024 for Llama-3.1-8B-Instruct ($d_{model}=4096$, $d_{inter}=14336$). MC and DC refer to \MC and \DC respectively.}
\vspace{-1em}
\label{tab:flops_mem_sparsity_time}
\end{table}

We first define a metric called \textbf{Comparative Influential Factor (CIF)} to measure how much influence $u$ (or $h$) has on the final decision of $s$, especially in cases where it overrides the other component. Analogously, for instance, $\textup{CIF}^k(u,\textup{alive})$ measures how often $u$ “rescues” a component that would otherwise have been pruned by $h$, allowing it to survive in $s$ due to its strong contribution. Formally, this is computed as:
\begin{equation}
\text{CIF}^k(u,\text{alive}) = 
\frac{|S^k \land \neg H^k|}
     {|S^k|}
\end{equation}

This formulation follows from the definition of $s$ as the elementwise product of $u$ and $h$. When $s[i]$ is alive but $h[i]$ is small enough to be pruned, it implies that $u[i]$ must have been large enough to compensate, effectively “saving” that entry.

Next, we define the \textbf{Comparative Agreement Factor (CAF)} to evaluate how often one signal aligns with $s$ while the other disagrees. For instance, $\textup{CAF}^k(u,\textup{alive})$ measures how frequently $u$ agrees with $s$ on keeping a component, specifically when $h$ disagrees. This is given by:
\begin{equation}
\text{CAF}^k(u,\text{alive}) = 
\frac{|S^k \land \neg H^k \land U^k |}
     {|S^k|}
\end{equation}

Both CIF and CAF can also be defined symmetrically for the “dead” case by inverting the roles of activation and pruning. 
\begin{figure}[t]
    \centering
    \includegraphics[width=\linewidth]{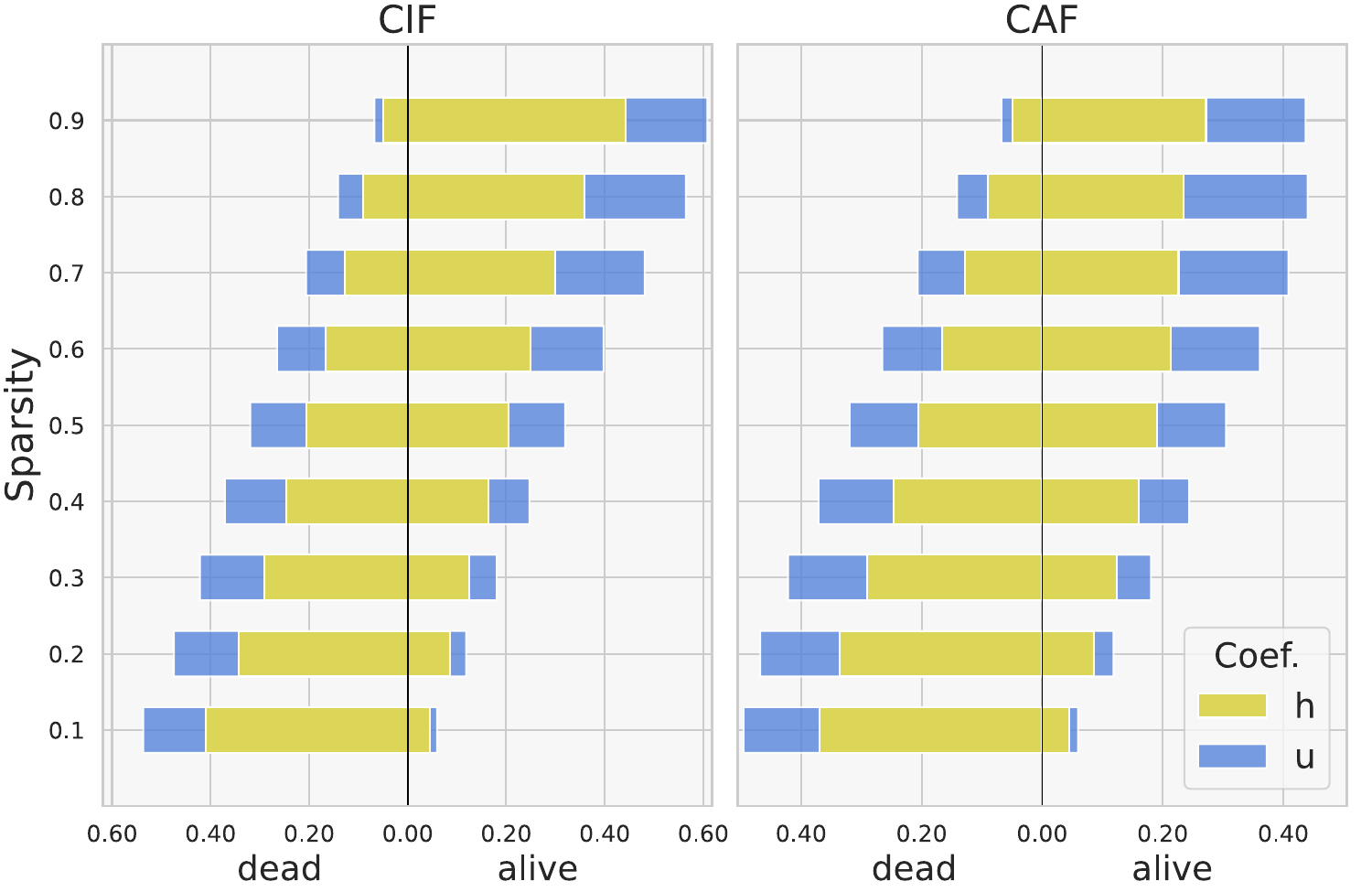}
    \caption{Tornado plots of CIF and CAF across whitening ratios. Bars to the right indicate the proportion of $\textup{CIF}^k(\cdot,alive)$, while those to the left indicate $\textup{CIF}^k(\cdot,dead)$.}
    \label{fig:CIF_and_CAF}
     \vspace{-1.5em}
\end{figure}

As shown in \autoref{fig:CIF_and_CAF}, $u$ outperforms $h$ across all sparsity levels in both CIF and CAF. These results suggest that $u$ more closely reflects the true activation behavior captured by $s$ and exerts a greater direct impact on sparsity decisions than $h$. In other words, $u$ is more effective at preserving important activations and filtering out unimportant ones, explaining \MC's stable and reliable performance under sparsity.

Nevertheless, since \MC still relies on an indirect coefficient $u$, it cannot fully match the upper-bound performance of \DC, which uses the full signal $s$ directly. Despite $u$'s strong CIF and CAF scores, substantial mismatches with respect to the oracle mask $S^k$ remain, with peak CIF values reaching only about 0.6 and CAF values about 0.4, underscoring the need for future work to translate \DC's upper-bound potential into \Spprac deployments.

\paragraph{Possible Predictor Candidate: TernaryLinear}
\label{para:alter_predictor}

\DC demonstrates a theoretically sound and effective sparse activation strategy, achieving strong performance in the \Spideal setting. However, in the \Spprac scenario, performance degradation occurs due to the predictor's limited accuracy in recovering optimal sparsity patterns. This reflects the difficulty of the prediction task rather than a flaw in the sparsity criterion itself. The task simultaneously demands precision and computational efficiency, presenting a significant challenge with considerable room for improvement.

To empirically explore this potential, we evaluate an alternative predictor architecture, \textbf{TernaryLinear}, whose weights are quantized as $\theta_{\text{ternary}} \in \{-1,\,0, +1\}^{d_{\text{model}} \times d_{\text{inter}}}$. We compare its performance with the previously utilized low-rank approximator. TernaryLinear achieves significant parameter compression by sacrificing numerical precision while preserving the matrix rank structure. Motivated by recent studies demonstrating successful LLM pretraining with ternary quantization while retaining strong model performance \citep{Ma2024-ow}, we regard TernaryLinear as a promising candidate due to its demonstrated expressiveness even under aggressive quantization.

As shown in \autoref{tab:ternary_verses_lowrank}, TernaryLinear outperforms the low-rank baseline in F1 score, while also being more compact in terms of memory footprint. This suggests that preserving rank information, even at the cost of numerical precision, is more effective for sparse mask recovery than the reverse approach.

However, TernaryLinear has not yet been adopted due to its relatively slower runtime despite its small size. This limitation stems not from algorithmic complexity, but rather the lack of optimized GPU kernel support for ultra-low-precision operations. Prior work \citep{ma2025bitnetb1582b4ttechnical} suggests that future advances in kernel optimization and ultra-low-bit quantization are needed to fully leverage such architectures. With these improvements, techniques like TernaryLinear could become viable candidates for enabling \DC to achieve its full \Spideal performance in \Spprac scenarios.

\begin{table}[t]
  \centering
  \small
  \begin{tabular}{lcc}
    \toprule
    Metric 
      & {TernaryLinear} 
      & {Low-Rank} \\
    \midrule
    Latency (ms) 
      & 0.082  
      & \textbf{0.030} \\
    Theoretical footprint (MiB) 
      & \textbf{112} 
      & 144 \\
    F1-score (\%) 
      & \textbf{0.435}
      & 0.403 \\
    \bottomrule
  \end{tabular}
  \caption{Comparison between TernaryLinear and the Low-Rank Approximator. Latency for TernaryLinear was measured using the BitBLAS library~\citep{ladder-osdi24}. F1 score is reported as the average binary classification performance on $S^{0.7}$ across all evaluated models.}
  \label{tab:ternary_verses_lowrank}
   \vspace{-1em}
\end{table}

\section{Conclusion}
We introduce \COUNTDOWN, a novel sparse activation framework for improving inference efficiency of large language models. To overcome the limitations of traditional non-linear activation-based sparsity, we reformulate the computation as a weighted sum over the FFN's down projection matrix, effectively capturing inherent sparsity in modern Gated-MLP architectures. From this perspective, we present two complementary strategies: \MC, which uses $u$ derived from a single matrix $W_{up}$ as its activation indicator, achieves faster inference and better performance preservation than prior state-of-the-art methods while remaining predictor-free. \DC directly leverages $s$, the coefficient vector of the weighted sum, for fine-grained sparsity selection, demonstrating robust performance even when skipping 90\% of computations under ideal conditions.

\section*{Limitations}
Like most prior work on sparse activation, our study assumes a single-batch greedy decoding setting in on-device environments. While this scenario is realistic for latency-sensitive edge inference, it may be less applicable in multi-batch or server-based deployments. In such cases, strategies such as computing the union of predicted index sets $\Idx$ across multiple samples could be explored. However, such an approach would require further investigation into how much parameter activation can be shared across inputs, a direction we leave for future work.

Additionally, our sparsity criteria rely exclusively on activation magnitude. This choice offers clear interpretability and aligns well with the weighted-sum perspective we adopt. Nevertheless, alternative sparsity metrics, such as those explored by \citep{Akhauri2024-dk}, remain an open research avenue. Expanding beyond simple magnitude-based thresholding could further enhance the performance of sparse activation methods.

\section*{Ethical Considerations}

We affirm adherence to the ACL Rolling Review (ARR) ethical guidelines, explicitly addressing potential risks and responsible research practices. This research focuses on optimizing computational efficiency in large language models (LLMs), aimed at reducing resource usage and consequently lowering environmental impact. We foresee no direct risks or potential harms to individuals or communities resulting from this work.

Comprehensive details regarding the ethical use of scientific artifacts, reproducibility of computational experiments, and related considerations are thoroughly documented in \autoref{sec:exp_details}.

\section*{Acknowledgments}
This research was supported by Basic Science Research Program through the National Research Foundation of Korea(NRF) funded by the Ministry of Education(No. 2340012238). This work was also supported by Institute of Information \& Communications Technology Planning \& Evaluation (IITP) grant funded by the Korea government (MSIT) (RS-2025-02214591, Development of an Innovative AI Agent for Worker-Friendly Autonomous Manufacturing),  (RS-2024-00460011, Climate and Environmental Data Platform for Enhancing Climate Technology Capabilities in the Anthropocene (CEDP)), and (RS2021-II211343, Artificial Intelligence Graduate School Program (Seoul National University)). This work was also supported by the National Research Foundation of Korea (NRF) grant funded by the Korea government (MSIT) (RS-2024-00407803, RS-2025-23523657). This work was also supported by the BK21 FOUR Program (Education and Research Center for Industrial Innovation Analytics) funded by the Ministry of Education, Korea (No. 4120240214912).

\bibliography{custom}

\appendix

\clearpage

\section{Experimental Details}
\label{sec:exp_details}

\subsection{Hyperparameters for Predictors}
\label{sec:hyperparam}
\begin{table}[H]
    \centering
    \adjustbox{max width=\linewidth}{
    \begin{tabular}{lcc}
        \toprule
        Name & Low-Rank$\bigl(\theta_\text{A}$, $\theta_\text{B}\bigr)$ & BitLinear$\bigl(\theta_{\text{ternary}}\bigr)$ \\
        \midrule
        Loss & \multicolumn{2}{c}{Binary Cross Entropy} \\
        Sparsity ratio ($k$) & \multicolumn{2}{c}{$\{0.7, 0.8, 0.9\}$} \\
        Learning rate & \multicolumn{2}{c}{\{\textbf{1e-3}, 5e-4\}} \\
        Training batch size & \multicolumn{2}{c}{$\{\textbf{16}, 32\}$} \\
        Optimizer & \multicolumn{2}{c}{AdamW} \\
        Target & \multicolumn{2}{c}{Binary mask $s_{\text{alive}}^k$} \\
        Epochs & \multicolumn{2}{c}{$\{10, 20, 40, \textbf{80}\}$} \\
        Seed & \multicolumn{2}{c}{$42$} \\
        \midrule
        Predictor shape & Low-Rank Approximator & TernaryLinear \\
        $d_{\text{rank}}$ & $\{128, 256, \textbf{512}, 1024\}$ & -- \\
        \midrule
        Hardware & \multicolumn{2}{c}{1 $\times$ NVIDIA A100 80GB} \\
        \bottomrule
    \end{tabular}
    }
    \caption{Hyperparameter settings and additional reproducibility details for training predictors used in \DC. All experiments were conducted using a single run without multiple random seeds. }
    \label{tab:hyperparams}
    \vspace{-0.5em}
\end{table}

\subsection{Environments}
\label{para:exp_env}
All experiments were performed on an NVIDIA A100 80GB GPU. We used Triton v3.1.0 for custom kernel development, while the rest of the experimental pipeline was built on HuggingFace Transformers v4.51.3, PyTorch v2.5.1, and CUDA v12.1.

\subsection{Dataset Description}

Table~\ref{tab:dataset_statistics} summarizes the licenses and dataset statistics used for evaluation.

We evaluate seven Natural Language Understanding(NLU) tasks and one Natural Language Generation(NLG) task focused on mathematical reasoning (GSM8K). All datasets primarily contain English text.

\begin{table}[h]
\centering
\resizebox{\linewidth}{!}{
\begin{tabular}{lccc}
\toprule
\textbf{Dataset} & \textbf{License} & \textbf{Train} & \textbf{Test} \\
\midrule
ARC-Easy & cc-by-sa-4.0 & 2251 (500) & 2376 \\
ARC-Challenge & cc-by-sa-4.0 & 1119 (500) & 1172 \\
HellaSwag & MIT & 39905 (500) & 10042 \\
PIQA & AFL 3.0 & 16113 (500) & 1838 \\
WinoGrande & apache-2.0 & 40398 (500) & 1267 \\
OpenBookQA & apache-2.0 & 4957 (500) & 500 \\
TruthfulQA & apache-2.0 & 0 & 817 \\
GSM8K & MIT & 0 & 1319 \\
\bottomrule
\end{tabular}}
\caption{Summary statistics and licenses for datasets used in evaluation. Following previous research \citep{Akhauri2024-dk}, we used subsets of each downstream task's training set, each containing 500 examples.}
\label{tab:dataset_statistics}
\end{table}

\newpage

\section{Pseudo Codes}
\subsection{Training Procedures}
\DC's predictor training only requires 2 hours for smaller models (Llama-3.1-8B, Gemma-2-9B) and 4 hours for larger models (Phi-4, Qwen2.5-14B) on a single NVIDIA A100 GPU. The resulting predictors are as compact as 576-900MB, representing merely 6-7\% of model parameters.

\begin{algorithm}[!ht]
\footnotesize
\KwIn{Training samples $\{x_j\}_{j=1}^{N}$, Target module $GatedMLP$, Target sparsity ratio $k$}
\KwOut{Trained predictor parameters $\theta$}

\ForEach{training sample $x_j$}{
    $s_j \leftarrow$ Compute $GatedMLP(x_j)$\;
    $s_j \leftarrow$ Binarize using $\text{Quantile}(k, |s_j|)$\;
}
\If{Predictor is Low-Rank}{
    Initialize parameters $\theta_\text{A}$, $\theta_\text{B}$\;
}
\ElseIf{Predictor is TernaryLinear}{
    Initialize parameters $\theta_{\text{ternary}}$\;
}
\ForEach{training iteration}{
    Sample mini-batch $\{x_b, s_b\}$\;
    \If{Predictor is Low-Rank}{
        $\hat{s}_b = x_b \cdot \theta_\text{A} \cdot \theta_\text{B}$\;
    }
    \ElseIf{Predictor is TernaryLinear}{
        $\hat{s}_b = x_b \cdot \theta_{\text{ternary}}$\;
    }
    Compute BCE loss between $\hat{s}_b$ and $s_b$\;
    Update predictor parameters $\theta$\;
}
\caption{Training the predictor for a Gated-MLP in \DC}
\label{algo:predictrain}
\end{algorithm}

\newpage

\subsection{Kernel in Detail: \MC}
\begin{algorithm}
\footnotesize
\caption{\MC Inference Kernel (Triton-style)}
\label{algo:mcountdown_kernel}
\KwIn{$X,\, W_{\text{up}},\,\hat{\tau}_\text{M} $}
\KwOut{$U, \text{Mask}$}
\textbf{\# PyTorch} \;
\vspace{0.3em}
$U \gets X\,@\,W_{\text{up}}$ \;
$\text{Mask} \gets (|U| \geq \hat{\tau}_\text{M})$ \;

\vspace{0.7em}

\KwIn{$X, U, W_{\text{gate}}, \text{Mask}, \text{BLK}_M, \text{BLK}_N$}
\KwOut{$S$}
\textbf{\# Triton 1} \;
\vspace{0.3em}
$\text{start\_m} \gets \text{tl.program\_id}(0)$ \;
$\text{rm} \gets \text{start\_m} \times \text{BLK}_M + \text{tl.arange}(0, \text{BLK}_M)$ \;
$\text{rn} \gets \text{tl.arange}(0, \text{BLK}_N)$ \;
$\text{Mask} \gets \text{Mask} + \text{rm}$ \;
$\text{flag} \gets \text{tl.load}(\text{Mask}) > 0$ \;
$W_{\text{gate}}\!\gets\!W_{\text{gate}}\!+\!(\text{rm}[:,\!\text{None}]\!\times \!d_{model}\!+\!\text{rn}[\text{None},\!:])$ \;
$X \gets X + \text{rn}$ \;
$\text{acc} \gets \text{tl.zeros}((\text{BLK}_M))$ \;
$\text{i\_mask} \gets \text{flag}[:, \text{None}]$ \;

\ForEach{block in rn}{
    $w \gets \text{tl.load}(W_{\text{gate}}, \text{mask}=\text{i\_mask}, \text{other}=0$ \;
    $x \gets \text{tl.load}(X)$ \;
    $\text{acc} \gets \text{acc} + \text{tl.sum}(w \times x[\text{None}, :], 1)$ \;
    $W_{\text{gate}} \gets W_{\text{gate}} + \text{BLK}_N$ \;
    $X \gets X + \text{BLK}_N$ \;
}
$U \gets U + \text{rm}$ \;
$u \gets \text{tl.load}(U, \text{mask}=\text{flag}, \text{other}=0)$ \;
$\text{acc} \gets \text{silu}(\text{acc}) \times u$ \;
$S \gets S + \text{rm}$ \;
$\text{tl.store}(S, \text{acc}, \text{mask}=rm < d_{inter})$ \;

\vspace{0.7em}
\KwIn{$S, W_{\text{down}}, \text{Mask}, \text{BLK}_M, \text{BLK}_N$}
\KwOut{$Y$}
\textbf{\# Triton 2} \;
\vspace{0.3em}
$\text{start\_m} \gets \text{tl.program\_id}(0)$ \;
$\text{start\_n} \gets \text{tl.program\_id}(1)$ \;
$\text{rm} \gets \text{start\_m} \times \text{BLK}_M + \text{tl.arange}(0, \text{BLK}_M)$ \;
$\text{rn} \gets \text{start\_n} \times \text{BLK}_N + \text{tl.arange}(0, \text{BLK}_N)$ \;
$\text{Mask} \gets \text{Mask} + \text{rm}$ \;
$\text{flag} \gets \text{tl.load}(\text{Mask}) > 0$ \;
$W_{\text{down}}\!\gets\!W_{\text{down}}\!+\!(\text{rm}[:,\!\text{None}]\!\times\!d_{model}\!+\!\text{rn}[\text{None},\!:])$ \;
$S \gets S + \text{rm}$ \;
$w\!\gets\!\text{tl.load}(W_{\text{down}},\!\text{mask}\!=\!\text{flag}[:,\!\text{None}],\!\text{other}\!=\!0)$ \;
$x \gets \text{tl.load}(S)$ \;
$\text{acc} \gets \text{tl.sum}(w \times x[:, \text{None}], 0)$ \;
$Y \gets Y + \text{rn}$ \;
$\text{tl.atomic\_add}(Y, \text{acc})$ \;

\end{algorithm}

\newpage

\subsection{Kernel in Detail: \DC}
\begin{algorithm}[!ht]
\footnotesize
\caption{\DC Inference Kernel (Triton-style)}
\label{algo:dcountdown_kernel}

\KwIn{$X, \theta_{\text{A}}, \theta_{\text{B}}, \tau_D$}
\KwOut{$\text{Mask}$}
\textbf{\# PyTorch} \;
\vspace{0.3em}
$\hat{s} \gets X\,@\,\theta_{\text{A}}\,@\,\theta_{\text{B}}$ \;
$\text{Mask} \gets (\hat{s}  \geq \tau_D)$ \;

\vspace{0.7em}

\KwIn{$X, W_{\text{gate}}, W_{\text{up}}, \text{Mask}, \text{BLK}_M, \text{BLK}_N$}
\KwOut{$S$}

\textbf{\# Triton 1} \;
\vspace{0.3em}
$\text{start\_m} \gets \text{tl.program\_id}(0)$ \;
$\text{rm} \gets \text{start\_m} \times \text{BLK}_M + \text{tl.arange}(0, \text{BLK}_M)$ \;
$\text{rn} \gets \text{tl.arange}(0, \text{BLK}_N)$ \;
$\text{Mask} \gets \text{Mask} + \text{rm}$ \;
$\text{flag} \gets \text{tl.load}(\text{Mask}) > 0$ \;
$W_{\text{gate}}\!\gets\!W_{\text{gate}}\!+\!(\text{rm}[:,\!\text{None}]\!\times \!d_{model}\!+\!\text{rn}[\text{None},\!:])$ \;
$W_{\text{up}}\!\gets\!W_{\text{up}}\!+\!(\text{rm}[:,\!\text{None}]\!\times \!d_{model}\!+\!\text{rn}[\text{None},\!:])$ \;
$X \gets X + \text{rn}$ \;
$\text{gate} \gets \text{tl.zeros}([\text{BLK}_M])$ \;
$\text{up} \gets \text{tl.zeros}([\text{BLK}_M])$ \;
$\text{i\_mask} \gets \text{flag}[:, \text{None}]$ \;
\ForEach{block in rn}{
    $w_{\text{gate}} \gets \text{tl.load}(W_{\text{gate}}, \text{mask}=\text{i\_mask}, \text{other}=0)$ \;
    $w_{\text{up}} \gets \text{tl.load}(W_{\text{up}}, \text{mask}=\text{i\_mask}, \text{other}=0)$ \;
    $x \gets \text{tl.load}(X)$ \;
    $\text{gate} \gets \text{gate} + \text{tl.sum}(w_{\text{gate}} \times x[\text{None}, :], \text{axis}=1)$ \;
    $\text{up} \gets \text{up} + \text{tl.sum}(w_{\text{up}} \times x[\text{None}, :], \text{axis}=1)$ \;
    $\text{X} \gets \text{X} + \text{BLK}_N$ \;
    $W_{\text{gate}} \gets W_{\text{gate}} + \text{BLK}_N$ \;
    $W_{\text{up}} \gets W_{\text{up}} + \text{BLK}_N$ \;
}

$\text{up} \gets \text{up} \times \text{SiLU}(\text{gate})$ \;
$\text{tl.store}(S, \text{up}, \text{mask} = \text{rm} < M)$ \;

\vspace{0.7em}
\KwIn{$S, W_{\text{down}}, \text{Mask}, \text{BLK}_M, \text{BLK}_N$}
\KwOut{$Y$}
\textbf{\# Triton 2} \;
\vspace{0.3em}
$\text{start\_m} \gets \text{tl.program\_id}(0)$ \;
$\text{start\_n} \gets \text{tl.program\_id}(1)$ \;
$\text{rm} \gets \text{start\_m} \times \text{BLK}_M + \text{tl.arange}(0, \text{BLK}_M)$ \;
$\text{rn} \gets \text{start\_n} \times \text{BLK}_N + \text{tl.arange}(0, \text{BLK}_N)$ \;
$\text{Mask} \gets \text{Mask} + \text{rm}$ \;
$\text{flag} \gets \text{tl.load}(\text{Mask}) > 0$ \;
$W_{\text{down}}\!\gets\!W_{\text{down}}\!+\!(\text{rm}[:,\!\text{None}]\!\times\!d_{model}\!+\!\text{rn}[\text{None},\!:])$ \;
$S \gets S + \text{rm}$ \;
$w\!\gets\!\text{tl.load}(W_{\text{down}},\!\text{mask}\!=\!\text{flag}[:,\!\text{None}],\!\text{other}\!=\!0)$ \;
$x \gets \text{tl.load}(S)$ \;
$\text{acc} \gets \text{tl.sum}(w \times x[:, \text{None}], 0)$ \;
$Y \gets Y + \text{rn}$ \;
$\text{tl.atomic\_add}(Y, \text{acc})$ \;

\end{algorithm}

\section{Full Results}
\label{sec:full_result}
\begin{figure}
    \centering
    \includegraphics[width=\linewidth]{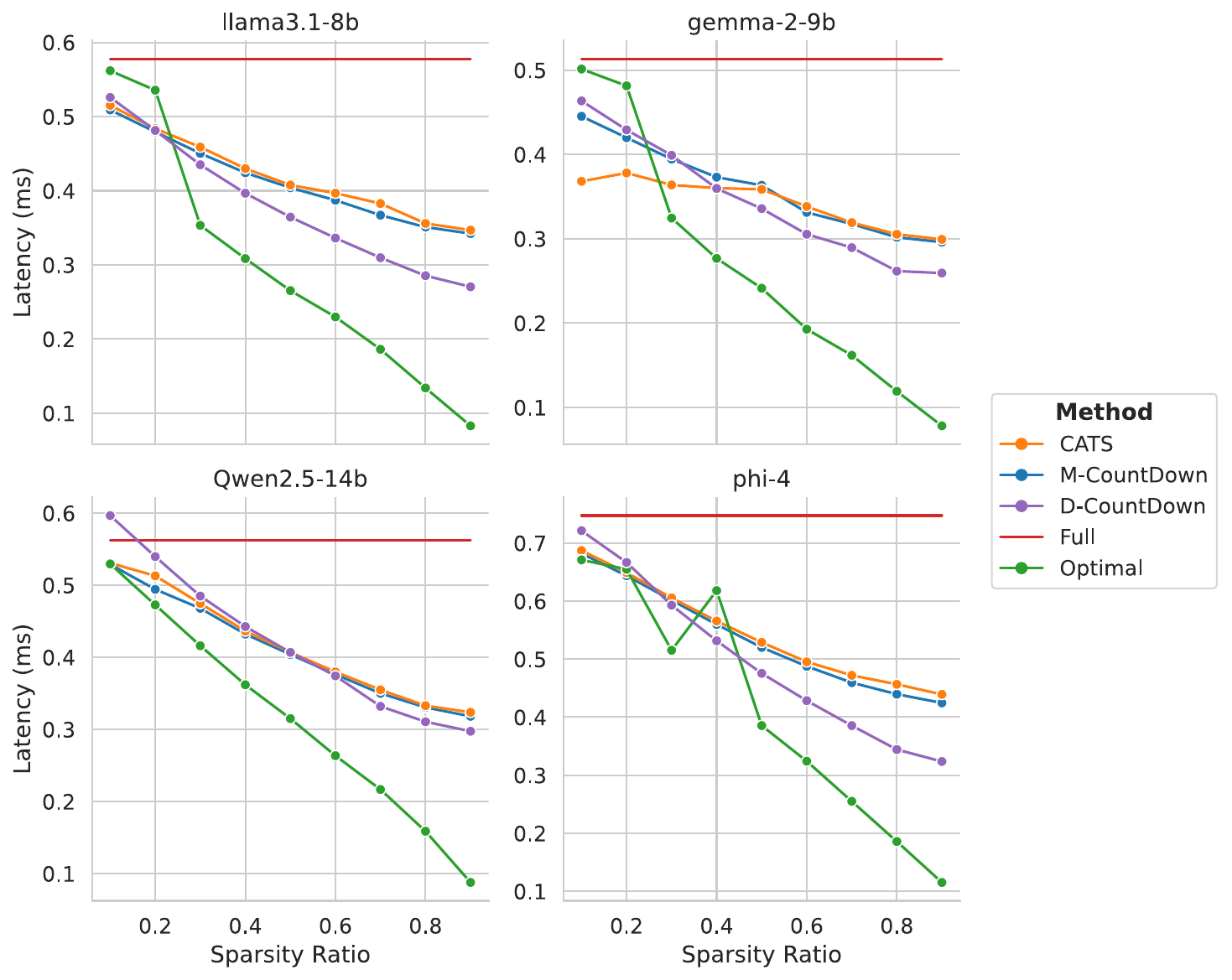}
    \caption{All results for kernel speed.}
    \label{fig:apdx_kernel}
    \vspace{-1em}
\end{figure}

All downstream task results are in \autoref{tab:full_taskwise_ideal} and \autoref{tab:full_taskwise_pred}. Chat performance results are shown in \autoref{tab:win_sparsity}. Kernel Speed results are shown in \autoref{fig:apdx_kernel}. Sparsity\textsuperscript{Real} indicates the actual proportion of indicator elements filtered out during \Spprac inferences.

\newpage

\begin{table}[t]
  \centering
  \begin{minipage}{\linewidth}
    \centering
    \resizebox{\linewidth}{!}{
    \begin{tabular}{ll|ccc|ccc}
      \toprule
      \multirow{3}{*}{\textbf{Method}} & \multirow{3}{*}{\textbf{Scenario}} 
        & \multicolumn{3}{c|}{\textbf{Llama-3.1-8B}} 
        & \multicolumn{3}{c}{\textbf{Gemma-2-9B}} \\
      \cmidrule(lr){3-5}\cmidrule(lr){6-8}
      & & \multicolumn{3}{c|}{Target Sparsity} & \multicolumn{3}{c}{Target Sparsity} \\
      & & 0.70 & 0.80 & 0.90 & 0.70 & 0.80 & 0.90 \\
      \midrule
      
      \multirow{3}{*}{CATS} & \Spideal 
        & 1.02 & 0.48 & 0.50 & 35.41 & 2.55 & 0.00 \\
      \cmidrule(lr){2-8}
      & \Spprac (Win)
        & 3.26 & 0.55 & 0.72 & 40.76 & 6.72 & 0.00 \\
      & \Spprac (Sparsity\textsuperscript{Real})
        & \textit{70.8} & \textit{80.0} & \textit{89.7} & \textit{68.8} & \textit{79.3} & \textit{88.1} \\
      \midrule
      
      \multirow{3}{*}{DC} & \Spideal
        & 45.79 & 39.33 & 11.85 & 50.44 & 48.90 & 37.79 \\
      \cmidrule(lr){2-8}
      & \Spprac (Win)
        & 1.35 & 1.57 & 0.77 & 6.72 & 7.99 & 3.57 \\
      & \Spprac (Sparsity\textsuperscript{Real})
        & \textit{68.8} & \textit{71.5} & \textit{80.8} & \textit{67.0} & \textit{72.9} & \textit{83.2} \\
      \midrule
      
      \multirow{3}{*}{MC} & \Spideal
        & 46.59 & 2.74 & 0.60 & 47.81 & 41.83 & 6.91 \\
      \cmidrule(lr){2-8}
      & \Spprac (Win)
        & 9.68 & 3.19 & 0.74 & 48.78 & 47.88 & 28.08 \\
      & \Spprac (Sparsity\textsuperscript{Real})
        & \textit{72.7} & \textit{82.3} & \textit{91.0} & \textit{68.0} & \textit{77.8} & \textit{87.6} \\
      \bottomrule
    \end{tabular}}
  \end{minipage}
  
  \vspace{0.5em}
  
  \begin{minipage}{\linewidth}
    \centering
    \resizebox{\linewidth}{!}{
    \begin{tabular}{ll|ccc|ccc}
      \toprule
      \multirow{3}{*}{\textbf{Method}} & \multirow{3}{*}{\textbf{Scenario}} 
        & \multicolumn{3}{c|}{\textbf{Qwen2.5-14B}} 
        & \multicolumn{3}{c}{\textbf{Phi-4}} \\
      \cmidrule(lr){3-5}\cmidrule(lr){6-8}
      & & \multicolumn{3}{c|}{Target Sparsity} & \multicolumn{3}{c}{Target Sparsity} \\
      & & 0.70 & 0.80 & 0.90 & 0.70 & 0.80 & 0.90 \\
      \midrule
      
      \multirow{3}{*}{CATS} & \Spideal 
        & 21.94 & 0.38 & 0.00 & 42.05 & 3.45 & 0.25 \\
      \cmidrule(lr){2-8}
      & \Spprac (Win)
        & 33.62 & 7.80 & 0.00 & 48.87 & 26.81 & 0.31 \\
      & \Spprac (Sparsity\textsuperscript{Real})
        & \textit{70.0} & \textit{80.0} & \textit{89.1} & \textit{67.6} & \textit{78.3} & \textit{89.7} \\
      \midrule
      
      \multirow{3}{*}{DC} & \Spideal
        & 50.10 & 48.71 & 32.73 & 49.10 & 46.46 & 37.45 \\
      \cmidrule(lr){2-8}
      & \Spprac (Win)
        & 4.40 & 0.77 & 0.20 & 1.11 & 0.90 & 0.12 \\
      & \Spprac (Sparsity\textsuperscript{Real})
        & \textit{66.6} & \textit{82.8} & \textit{87.6} & \textit{65.4} & \textit{78.5} & \textit{86.8} \\
      \midrule
      
      \multirow{3}{*}{MC} & \Spideal
        & 45.01 & 36.90 & 2.83 & 43.96 & 35.43 & 5.28 \\
      \cmidrule(lr){2-8}
      & \Spprac (Win)
        & 48.57 & 42.51 & 15.89 & 46.24 & 41.62 & 18.82 \\
      & \Spprac (Sparsity\textsuperscript{Real})
        & \textit{70.0} & \textit{80.0} & \textit{90.0} & \textit{70.1} & \textit{79.6} & \textit{89.6} \\
      \bottomrule
    \end{tabular}}
  \end{minipage}
  
  \caption{Summary of Win Rate on AlpacaEval 2.0}
  \label{tab:win_sparsity}
\end{table}

\section{Theoretical Analysis Details}
\label{sec:calc_explain}

\subsection{Notation}
\begin{table}[H]
\centering
\resizebox{0.8\linewidth}{!}{
\begin{tabular}{@{}ll@{}}
\toprule
\textbf{Notation} & \textbf{Explanation} \\
\midrule
\(d_m\)               & $d_{model}$ \\
\(d_i\)               & $d_{inter}$ \\
\(d_r\)               & $d_{rank}$ \\
\(s\)                 & $int({d_{inter}} \times k)$ \\
\(c_{\mathrm{act}}\)  & act FLOPs (e.g.\ SiLU \(\approx5\)) \\
\bottomrule
\end{tabular}}

\caption{Notation Used in Theoretical Analysis}
\vspace{-1em}
\label{tab:notation}
\end{table}

\newpage

\subsection{Theoretical FLOPs Analysis}

\begin{table}[H]
\centering
\resizebox{0.85\linewidth}{!}{
\begin{tabular}{@{}l l l@{}}
\toprule
\textbf{Method} & \textbf{Compute} & \textbf{Explanation} \\
\midrule
\makecell[l]{Dense}
  & $\displaystyle\begin{array}{@{}l@{}}
      6\,d_m\,d_i \\
      +\,c_{\mathrm{act}}\,d_i \\
      +\,d_i
    \end{array}$
  & $\displaystyle\begin{array}{@{}l@{}}
      \text{Full GEMV}\times3 \\
      \text{Full }\sigma \\
      \text{Full }\odot
    \end{array}$
  \\
\midrule
\makecell[l]{CATS}
  & $\displaystyle\begin{array}{@{}l@{}}
      2\,d_m\,d_i \\
      +\;c_{\mathrm{act}}\,d_i \\
      +\;2\,d_i \\
      +\;2\,d_m\,s \\
      +\;s \\
      +\;2\,d_m\,s
    \end{array}$
  & $\displaystyle\begin{array}{@{}l@{}}
      \text{Full GEMV }W_{gate} \\
      \text{Full }\sigma \\
      \text{Apply abs and }\Thld \\
      \text{Sparse GEMV }W_{up} \\
      \text{Sparse }\odot \\
      \text{Sparse GEMV }W_{down}
    \end{array}$
  \\
\midrule
\makecell[l]{\MC}
  & $\displaystyle\begin{array}{@{}l@{}}
      2\,d_m\,d_i \\
      +\;2\,d_i \\
      +\;2\,d_m\,s \\
      +\;c_{\mathrm{act}}\,s \\
      +\;s \\
      +\;2\,d_m\,s
    \end{array}$
  & $\displaystyle\begin{array}{@{}l@{}}
      \text{Full GEMV }W_{up} \\
      \text{Apply abs and }\Thld \\
      \text{Sparse GEMV }W_{gate} \\
      \text{Sparse }\sigma \\
      \text{Sparse }\odot \\
      \text{Sparse GEMV }W_{down}
    \end{array}$
  \\
\midrule
\makecell[l]{\DC}
  & $\displaystyle\begin{array}{@{}l@{}}
      2\,d_m\,d_r \\
      +\;2\,d_r\,d_i \\
      +\;d_i \\
      +\;4\,d_m\,s \\
      +\;c_{\mathrm{act}}\,s \\
      +\;s \\
      +\;2\,d_m\,s
    \end{array}$
  & $\displaystyle\begin{array}{@{}l@{}}
      \text{Low-rank GEMV }\theta_A \\
      \text{Low-rank GEMV }\theta_B \\
      \text{Apply }\Thld \\
      \text{Sparse GEMV }W_{gate},W_{up} \\
      \text{Sparse }\sigma \\
      \text{Sparse }\odot \\
      \text{Sparse GEMV }W_{down}
    \end{array}$
  \\
\bottomrule
\end{tabular}}
\caption{Comparison of Theoretical FLOPs Across Methods}
\label{tab:flops-comparison}
\end{table}

\subsection{Theoretical Memory Traffic Analysis}
{\captionsetup{labelformat=empty}
\begin{table}[h]
\centering
\resizebox{0.85\linewidth}{!}{
\begin{tabular}{@{}l l l@{}}
\toprule
\textbf{Method} & \textbf{Mem. R/W} & \textbf{Explanation} \\
\midrule
\makecell[l]{Dense}
  & $\displaystyle\begin{array}{@{}l@{}}
      2\,d_m\,d_i \\
      +\,2\,d_m \\
      +\,2\,d_i \\
      +\,d_i \\
      +\,d_i \\
      +\,2\,d_i \\
      +\,d_i \\
      +\,d_m\,d_i \\
      +\,d_i \\
      +\,d_m
    \end{array}$
  & $\displaystyle\begin{array}{@{}l@{}}
      \text{Read Full }W_{up},W_{gate} \\
      \text{Read }x\times2 \\
      \text{Write gate, up} \\
      \text{Read gate} \\
      \text{Write act\_gate} \\
      \text{Read act\_gate, up} \\
      \text{Write inter} \\
      \text{Read Full }W_{down} \\
      \text{Read inter} \\
      \text{Write }y
    \end{array}$
  \\
\midrule
\makecell[l]{CATS}
  & $\displaystyle\begin{array}{@{}l@{}}
      d_m\,d_i \\
      +\,d_m \\
      +\,d_i \\
      +\,d_i \\
      +\,d_i \\
      +\,d_i \\
      +\,d_i \\
      +\,d_i \\
      +\,d_i \\
      +\,d_m\,s \\
      +\,d_m \\
      +\,s \\
      +\,d_i \\
      +\,d_i \\
      +\,d_m\,s \\
      +\,d_i \\
      +\,d_m
    \end{array}$
  & $\displaystyle\begin{array}{@{}l@{}}
      \text{Read Full }W_{gate} \\
      \text{Read }x \\
      \text{Write gate} \\
      \text{Read gate} \\
      \text{Write act\_gate} \\
      \text{Read act\_gate} \\
      \text{Write abs\_act\_gate} \\
      \text{Read abs\_act\_gate} \\
      \text{Write mask} \\
      \text{Read Sparse }W_{up} \\
      \text{Read }x \\
      \text{Read Sparse act\_gate} \\
      \text{Read mask} \\
      \text{Write inter} \\
      \text{Read Sparse }W_{down} \\
      \text{Read inter} \\
      \text{Write }y
    \end{array}$
    \\
\bottomrule
\end{tabular}}
\label{tab:memory-comparison-half}
\end{table}}

\newpage

\begin{table}[H]
\centering
\resizebox{\linewidth}{!}{
\begin{tabular}{@{}l l l@{}}
\toprule
\textbf{Method} & \textbf{Mem. R/W} & \textbf{Explanation} \\
\midrule
\makecell[l]{M-Countdown}
  & $\displaystyle\begin{array}{@{}l@{}}
      d_m\,d_i \\
      +\,d_m \\
      +\,d_i \\
      +\,d_i \\
      +\,d_i \\
      +\,d_i \\
      +\,d_i \\
      +\,d_m\,s \\
      +\,d_m \\
      +\,s \\
      +\,d_i \\
      +\,d_i \\
      +\,d_m\,s \\
      +\,d_i \\
      +\,d_m
    \end{array}$
  & $\displaystyle\begin{array}{@{}l@{}}
      \text{Read Full }W_{up} \\
      \text{Read }x \\
      \text{Write up} \\
      \text{Read up} \\
      \text{Write abs\_up} \\
      \text{Read abs\_up} \\
      \text{Write mask} \\
      \text{Read Sparse }W_{gate} \\
      \text{Read }x \\
      \text{Read Sparse up} \\
      \text{Read mask} \\
      \text{Write inter} \\
      \text{Read Sparse }W_{down} \\
      \text{Read inter} \\
      \text{Write }y
    \end{array}$
  \\
\midrule
\makecell[l]{D-Countdown}
  & $\displaystyle\begin{array}{@{}l@{}}
      d_m\,d_r \\
      +\,d_m \\
      +\,d_r \\
      +\,d_r\,d_i \\
      +\,d_r \\
      +\,d_i \\
      +\,d_i \\
      +\,d_i \\
      +\,2\,d_m\,s \\
      +\,d_m \\
      +\,d_i \\
      +\,d_i \\
      +\,d_m\,s \\
      +\,d_i \\
      +\,d_m
    \end{array}$
  & $\displaystyle\begin{array}{@{}l@{}}
      \text{Read }\theta_{A} \\
      \text{Read }x \\
      \text{Write latent} \\
      \text{Read }\theta_{B} \\
      \text{Read latent} \\
      \text{Write }\hat{s} \\
      \text{Read }\hat{s} \\
      \text{Write mask} \\
      \text{Read Sparse }W_{up},W_{gate} \\
      \text{Read }x \\
      \text{Read mask} \\
      \text{Write inter} \\
      \text{Read Sparse }W_{down} \\
      \text{Read inter} \\
      \text{Write }y
    \end{array}$
  \\
\bottomrule
\end{tabular}}
\caption{Comparison of Theoretical Memory Traffic Across Methods}
\label{tab:memory-comparison}
\end{table}

\begin{table*}[t]
\centering
\resizebox {0.8\textwidth} {!}{
\begin{tabular}{lccccccccc}
\toprule
\textbf{Sparsity} & \textbf{Method} & \textbf{ARC-C} & \textbf{TFQA} & \textbf{HS} & \textbf{ARC-E} & \textbf{PIQA} & \textbf{WG} & \textbf{OBQA} & \textbf{GSM8K}\\
\midrule
\multicolumn{10}{c}{\textbf{Llama-3.1-8B-Instruct}} \\
\midrule
\multirow{1}{*}{0.0} & DENSE & \textbf{0.520} & \textbf{0.367} & \textbf{0.590} & \textbf{0.819} & \textbf{0.800} & \textbf{0.737} & \textbf{0.336} & \textbf{0.760} \\
\midrule
\multirow{4}{*}{0.7} & DEJAVU & 0.292 & 0.229 & 0.272 & 0.445 & 0.553 & 0.503 & 0.218 & 0.000 \\
 & CATS & 0.453 & 0.343 & 0.523 & 0.754 & 0.739 & 0.653 & 0.298 & 0.003 \\
 & M-COUNTDOWN & 0.493 & \textbf{0.372} & 0.568 & 0.784 & 0.776 & 0.695 & 0.330 & 0.544 \\
 & D-COUNTDOWN & \textbf{0.509} & 0.370 & \textbf{0.592} & \textbf{0.812} & \textbf{0.795} & \textbf{0.727} & \textbf{0.332} & \textbf{0.688} \\
\midrule
\multirow{4}{*}{0.8} & DEJAVU & 0.282 & 0.231 & 0.273 & 0.440 & 0.557 & 0.511 & 0.228 & 0.000 \\
 & CATS & 0.358 & 0.326 & 0.428 & 0.651 & 0.676 & 0.582 & 0.278 & 0.000 \\
 & M-COUNTDOWN & 0.458 & 0.343 & 0.534 & 0.759 & 0.748 & 0.661 & 0.314 & 0.288 \\
 & D-COUNTDOWN & \textbf{0.502} & \textbf{0.356} & \textbf{0.585} & \textbf{0.809} & \textbf{0.789} & \textbf{0.713} & \textbf{0.334} & \textbf{0.605} \\
\midrule
\multirow{4}{*}{0.9} & DEJAVU & 0.296 & 0.230 & 0.273 & 0.455 & 0.557 & 0.530 & 0.236 & 0.000 \\
 & CATS & 0.293 & 0.252 & 0.303 & 0.495 & 0.574 & 0.534 & 0.242 & 0.000 \\
 & M-COUNTDOWN & 0.411 & 0.304 & 0.430 & 0.649 & 0.676 & 0.613 & 0.286 & 0.001 \\
 & D-COUNTDOWN & \textbf{0.484} & \textbf{0.330} & \textbf{0.548} & \textbf{0.776} & \textbf{0.755} & \textbf{0.680} & \textbf{0.312} & \textbf{0.313} \\
\midrule
\multicolumn{10}{c}{\textbf{Qwen2.5-14B-Instruct}} \\
\midrule
\multirow{1}{*}{0.0} & DENSE & \textbf{0.608} & \textbf{0.517} & \textbf{0.657} & \textbf{0.861} & \textbf{0.817} & \textbf{0.758} & \textbf{0.364} & \textbf{0.807} \\
\midrule
\multirow{4}{*}{0.7} & DEJAVU & 0.336 & 0.318 & 0.365 & 0.612 & 0.616 & 0.533 & 0.254 & 0.000 \\
 & CATS & 0.488 & 0.443 & 0.585 & 0.777 & 0.729 & 0.629 & 0.318 & 0.043 \\
 & M-COUNTDOWN & 0.573 & 0.488 & 0.638 & 0.829 & 0.792 & 0.704 & 0.352 & \textbf{0.776} \\
 & D-COUNTDOWN & \textbf{0.588} & \textbf{0.518} & \textbf{0.654} & \textbf{0.850} & \textbf{0.801} & \textbf{0.736} & \textbf{0.364} & 0.770 \\
\midrule
\multirow{4}{*}{0.8} & DEJAVU & 0.340 & 0.322 & 0.357 & 0.609 & 0.619 & 0.554 & 0.258 & 0.000 \\
 & CATS & 0.410 & 0.371 & 0.472 & 0.683 & 0.632 & 0.568 & 0.284 & 0.000 \\
 & M-COUNTDOWN & 0.532 & 0.476 & 0.614 & 0.813 & 0.743 & 0.670 & 0.352 & 0.681 \\
 & D-COUNTDOWN & \textbf{0.579} & \textbf{0.488} & \textbf{0.644} & \textbf{0.837} & \textbf{0.799} & \textbf{0.716} & \textbf{0.360} & \textbf{0.751} \\
\midrule
\multirow{4}{*}{0.9} & DEJAVU & 0.358 & 0.333 & 0.369 & 0.612 & 0.621 & 0.531 & 0.256 & 0.000 \\
 & CATS & 0.356 & 0.327 & 0.385 & 0.619 & 0.621 & 0.547 & 0.260 & 0.000 \\
 & M-COUNTDOWN & 0.468 & 0.421 & 0.525 & 0.736 & 0.686 & 0.589 & 0.304 & 0.100 \\
 & D-COUNTDOWN & \textbf{0.512} & \textbf{0.436} & \textbf{0.607} & \textbf{0.801} & \textbf{0.756} & \textbf{0.648} & \textbf{0.312} & \textbf{0.371} \\
\midrule
\multicolumn{10}{c}{\textbf{gemma-2-9b-it}} \\
\midrule
\multirow{1}{*}{0.0} & DENSE & \textbf{0.632} & \textbf{0.433} & \textbf{0.597} & \textbf{0.856} & \textbf{0.812} & \textbf{0.761} & \textbf{0.404} & \textbf{0.663} \\
\midrule
\multirow{4}{*}{0.7} & DEJAVU & 0.339 & 0.246 & 0.300 & 0.596 & 0.590 & 0.532 & 0.276 & 0.000 \\
 & CATS & 0.575 & 0.412 & 0.559 & 0.840 & 0.755 & 0.680 & 0.348 & 0.565 \\
 & M-COUNTDOWN & 0.605 & \textbf{0.421} & 0.592 & 0.849 & 0.793 & 0.726 & 0.374 & 0.632 \\
 & D-COUNTDOWN & \textbf{0.626} & 0.417 & \textbf{0.600} & \textbf{0.854} & \textbf{0.800} & \textbf{0.750} & \textbf{0.384} & \textbf{0.649} \\
\midrule
\multirow{4}{*}{0.8} & DEJAVU & 0.346 & 0.246 & 0.296 & 0.599 & 0.581 & 0.548 & 0.262 & 0.000 \\
 & CATS & 0.490 & 0.366 & 0.486 & 0.788 & 0.696 & 0.604 & 0.328 & 0.105 \\
 & M-COUNTDOWN & 0.583 & 0.408 & 0.582 & 0.842 & 0.767 & 0.707 & 0.360 & 0.610 \\
 & D-COUNTDOWN & \textbf{0.604} & \textbf{0.421} & \textbf{0.599} & \textbf{0.851} & \textbf{0.796} & \textbf{0.728} & \textbf{0.374} & \textbf{0.624} \\
\midrule
\multirow{4}{*}{0.9} & DEJAVU & 0.356 & 0.246 & 0.303 & 0.616 & 0.573 & 0.523 & 0.264 & 0.000 \\
 & CATS & 0.364 & 0.242 & 0.310 & 0.617 & 0.589 & 0.537 & 0.278 & 0.000 \\
 & M-COUNTDOWN & 0.534 & 0.383 & 0.517 & 0.799 & 0.727 & 0.648 & 0.344 & 0.438 \\
 & D-COUNTDOWN & \textbf{0.578} & \textbf{0.410} & \textbf{0.572} & \textbf{0.833} & \textbf{0.777} & \textbf{0.676} & \textbf{0.352} & \textbf{0.524} \\
\midrule
\multicolumn{10}{c}{\textbf{phi-4}} \\
\midrule
\multirow{1}{*}{0.0} & DENSE & \textbf{0.558} & \textbf{0.404} & \textbf{0.632} & \textbf{0.814} & \textbf{0.808} & \textbf{0.766} & \textbf{0.338} & \textbf{0.923} \\
\midrule
\multirow{4}{*}{0.7} & DEJAVU & 0.387 & 0.311 & 0.348 & 0.655 & 0.626 & 0.587 & 0.266 & 0.000 \\
 & CATS & 0.536 & 0.400 & 0.595 & 0.794 & 0.791 & 0.696 & 0.304 & 0.807 \\
 & M-COUNTDOWN & 0.533 & 0.384 & 0.616 & 0.800 & 0.796 & 0.733 & \textbf{0.334} & 0.888 \\
 & D-COUNTDOWN & \textbf{0.554} & \textbf{0.411} & \textbf{0.630} & \textbf{0.809} & \textbf{0.807} & \textbf{0.752} & 0.332 & \textbf{0.916} \\
\midrule
\multirow{4}{*}{0.8} & DEJAVU & 0.409 & 0.333 & 0.354 & 0.655 & 0.632 & 0.585 & 0.270 & 0.000 \\
 & CATS & 0.516 & 0.397 & 0.539 & 0.771 & 0.760 & 0.644 & 0.298 & 0.351 \\
 & M-COUNTDOWN & 0.503 & 0.386 & 0.594 & 0.792 & 0.778 & 0.715 & 0.330 & 0.767 \\
 & D-COUNTDOWN & \textbf{0.552} & \textbf{0.408} & \textbf{0.622} & \textbf{0.807} & \textbf{0.810} & \textbf{0.755} & \textbf{0.340} & \textbf{0.898} \\
\midrule
\multirow{4}{*}{0.9} & DEJAVU & 0.392 & 0.317 & 0.357 & 0.640 & 0.630 & 0.566 & 0.266 & 0.000 \\
 & CATS & 0.426 & 0.356 & 0.414 & 0.676 & 0.672 & 0.591 & 0.280 & 0.000 \\
 & M-COUNTDOWN & 0.479 & 0.370 & 0.524 & 0.759 & 0.728 & 0.654 & 0.296 & 0.287 \\
 & D-COUNTDOWN & \textbf{0.529} & \textbf{0.399} & \textbf{0.601} & \textbf{0.798} & \textbf{0.789} & \textbf{0.695} & \textbf{0.318} & \textbf{0.827} \\
\bottomrule
\end{tabular}}
\vspace{-0.5em}
\caption{\centering \Spideal scores compared to \Dense across all downstream tasks. \Dense scores are in bold, as well as the highest score for each task within each sparsity level.}
\label{tab:full_taskwise_ideal}
\end{table*}

\begin{table*}[t]
\centering
\resizebox {\textwidth} {!}{
\begin{tabular}{lcccccccccc}
\toprule
\textbf{Sparsity} & \textbf{Method} & \textbf{Sparsity}\textsuperscript{Real} & \textbf{ARC-C} & \textbf{TFQA} & \textbf{HS} & \textbf{ARC-E} & \textbf{PIQA} & \textbf{WG} & \textbf{OBQA} & \textbf{GSM8K}\\
\midrule
\multicolumn{11}{c}{\textbf{Llama-3.1-8B-Instruct}} \\
\midrule
\multirow{3}{*}{0.7}
 & CATS & 0.684 & 0.461 & 0.355 & 0.549 & 0.778 & 0.764 & 0.683 & \textbf{0.316} & 0.127 \\
 & M-COUNTDOWN & 0.709 & \textbf{0.484} & \textbf{0.375} & \textbf{0.574} & \textbf{0.788} & \textbf{0.778} & \textbf{0.708} & 0.310 & \textbf{0.547} \\
 & D-COUNTDOWN & 0.705 & 0.422 & 0.318 & 0.373 & 0.748 & 0.714 & 0.663 & 0.298 & 0.002 \\
\midrule
\multirow{3}{*}{0.8}
 & CATS & 0.784 & 0.420 & 0.322 & 0.495 & 0.718 & 0.721 & 0.624 & 0.296 & 0.000 \\
 & M-COUNTDOWN & 0.806 & \textbf{0.460} & \textbf{0.361} & \textbf{0.549} & \textbf{0.770} & \textbf{0.757} & \textbf{0.680} & \textbf{0.322} & \textbf{0.322} \\
 & D-COUNTDOWN & 0.739 & 0.382 & 0.306 & 0.388 & 0.688 & 0.673 & 0.621 & 0.292 & 0.003 \\
\midrule
\multirow{3}{*}{0.9}
 & CATS & 0.902 & 0.299 & 0.273 & 0.323 & 0.521 & 0.607 & 0.537 & 0.238 & 0.000 \\
 & M-COUNTDOWN & 0.895 & \textbf{0.416} & \textbf{0.321} & \textbf{0.471} & \textbf{0.711} & \textbf{0.721} & \textbf{0.620} & \textbf{0.304} & \textbf{0.009} \\
 & D-COUNTDOWN & 0.843 & 0.349 & 0.285 & 0.345 & 0.628 & 0.635 & 0.593 & 0.260 & 0.000 \\
\midrule
\multicolumn{11}{c}{\textbf{Qwen2.5-14B-Instruct}} \\
\midrule
\multirow{3}{*}{0.7}
 & CATS & 0.698 & 0.518 & 0.460 & 0.612 & 0.805 & 0.761 & 0.660 & 0.336 & 0.293 \\
 & M-COUNTDOWN & 0.719 & \textbf{0.590} & \textbf{0.509} & \textbf{0.640} & \textbf{0.838} & \textbf{0.792} & \textbf{0.712} & \textbf{0.358} & \textbf{0.767} \\
 & D-COUNTDOWN & 0.678 & 0.513 & 0.426 & 0.536 & 0.798 & 0.748 & 0.668 & 0.322 & 0.197 \\
\midrule
\multirow{3}{*}{0.8}
 & CATS & 0.802 & 0.472 & 0.421 & 0.551 & 0.754 & 0.712 & 0.627 & 0.284 & 0.000 \\
 & M-COUNTDOWN & 0.804 & \textbf{0.553} & \textbf{0.492} & \textbf{0.625} & \textbf{0.826} & \textbf{0.769} & \textbf{0.669} & \textbf{0.354} & \textbf{0.704} \\
 & D-COUNTDOWN & 0.827 & 0.454 & 0.394 & 0.468 & 0.740 & 0.693 & 0.615 & 0.292 & 0.000 \\
\midrule
\multirow{3}{*}{0.9}
 & CATS & 0.906 & 0.347 & 0.350 & 0.393 & 0.631 & 0.616 & 0.527 & 0.258 & 0.000 \\
 & M-COUNTDOWN & 0.889 & \textbf{0.492} & \textbf{0.450} & \textbf{0.580} & \textbf{0.794} & \textbf{0.727} & \textbf{0.632} & \textbf{0.320} & \textbf{0.287} \\
 & D-COUNTDOWN & 0.893 & 0.434 & 0.384 & 0.429 & 0.689 & 0.669 & 0.605 & 0.282 & 0.000 \\
\midrule
\multicolumn{11}{c}{\textbf{gemma-2-9b-it}} \\
\midrule
\multirow{3}{*}{0.7}
 & CATS & 0.695 & 0.580 & 0.427 & 0.567 & 0.843 & 0.770 & 0.693 & 0.368 & 0.593 \\
 & M-COUNTDOWN & 0.685 & \textbf{0.608} & \textbf{0.431} & \textbf{0.598} & \textbf{0.854} & \textbf{0.801} & \textbf{0.745} & \textbf{0.386} & \textbf{0.633} \\
 & D-COUNTDOWN & 0.689 & 0.567 & 0.403 & 0.493 & 0.821 & 0.751 & 0.702 & 0.364 & 0.340 \\
\midrule
\multirow{3}{*}{0.8}
 & CATS & 0.806 & 0.542 & 0.392 & 0.501 & 0.811 & 0.729 & 0.615 & 0.346 & 0.083 \\
 & M-COUNTDOWN & 0.779 & \textbf{0.596} & \textbf{0.412} & \textbf{0.589} & \textbf{0.847} & \textbf{0.788} & \textbf{0.712} & 0.370 & \textbf{0.618} \\
 & D-COUNTDOWN & 0.755 & 0.564 & 0.401 & 0.506 & 0.819 & 0.758 & 0.702 & \textbf{0.374} & 0.381 \\
\midrule
\multirow{3}{*}{0.9}
 & CATS & 0.911 & 0.340 & 0.258 & 0.306 & 0.617 & 0.586 & 0.514 & 0.262 & 0.000 \\
 & M-COUNTDOWN & 0.875 & \textbf{0.573} & \textbf{0.395} & \textbf{0.554} & \textbf{0.829} & \textbf{0.761} & \textbf{0.686} & \textbf{0.360} & \textbf{0.544} \\
 & D-COUNTDOWN & 0.853 & 0.529 & 0.383 & 0.492 & 0.806 & 0.747 & 0.665 & 0.354 & 0.187 \\
\midrule
\multicolumn{11}{c}{\textbf{phi-4}} \\
\midrule
\multirow{3}{*}{0.7}
 & CATS & 0.675 & 0.539 & \textbf{0.417} & 0.613 & 0.801 & 0.795 & 0.724 & 0.322 & 0.856 \\
 & M-COUNTDOWN & 0.707 & \textbf{0.540} & 0.393 & \textbf{0.620} & \textbf{0.804} & \textbf{0.796} & \textbf{0.736} & \textbf{0.332} & \textbf{0.894} \\
 & D-COUNTDOWN & 0.687 & 0.471 & 0.368 & 0.485 & 0.750 & 0.733 & 0.685 & 0.294 & 0.208 \\
\midrule
\multirow{3}{*}{0.8}
 & CATS & 0.771 & 0.525 & \textbf{0.390} & 0.587 & \textbf{0.795} & \textbf{0.786} & 0.673 & 0.300 & 0.675 \\
 & M-COUNTDOWN & 0.799 & \textbf{0.527} & 0.381 & \textbf{0.607} & 0.793 & 0.784 & \textbf{0.715} & \textbf{0.334} & \textbf{0.817} \\
 & D-COUNTDOWN & 0.815 & 0.418 & 0.343 & 0.438 & 0.707 & 0.692 & 0.657 & 0.268 & 0.036 \\
\midrule
\multirow{3}{*}{0.9}
 & CATS & 0.889 & 0.458 & 0.360 & 0.460 & 0.713 & 0.692 & 0.605 & 0.294 & 0.000 \\
 & M-COUNTDOWN & 0.894 & \textbf{0.498} & \textbf{0.386} & \textbf{0.563} & \textbf{0.777} & \textbf{0.750} & \textbf{0.698} & \textbf{0.316} & \textbf{0.450} \\
 & D-COUNTDOWN & 0.895 & 0.408 & 0.294 & 0.404 & 0.674 & 0.668 & 0.616 & 0.270 & 0.000 \\
\bottomrule
\end{tabular}}
\caption{\centering \Spprac scores compared across all downstream tasks. Bold indicates the highest score at each sparsity level for each task.}
\label{tab:full_taskwise_pred}
\end{table*}

\end{document}